\documentclass[runningheads]{llncs}
\usepackage[year=2026,ID=1424]{eccv}
\usepackage{eccvabbrv}

\usepackage{hyperref}

\usepackage{orcidlink}
\usepackage[utf8]{inputenc}
\usepackage[T1]{fontenc}
\usepackage[english]{babel}
\usepackage{fontawesome5}
\usepackage{tabularx}
\usepackage{lipsum}
\usepackage[table,xcdraw]{xcolor}
\usepackage{hhline}
\usepackage{bm}
\usepackage{pdfpages}
\usepackage{acro}
\usepackage{tocbibind}
\usepackage{bookmark}
\usepackage{subcaption}
\usepackage{etoolbox}
\usepackage{wrapfig}
\usepackage{stmaryrd}
\usepackage{rotating}
\usepackage{mwe}
\usepackage{multicol}
\usepackage{multirow}
\usepackage{lettrine}
\usepackage{tikz}
\usetikzlibrary{calc}
\usepackage{environ}
\usepackage{adjustbox}
\usepackage{siunitx}
\usepackage{pifont}
\usepackage{mathtools}
\usepackage{enumitem}
\usepackage{amsmath,amssymb}
\usepackage{mdframed}
\usepackage{pgfplots}
\pgfplotsset{compat = newest}
\usepackage{booktabs}
\usepackage{tcolorbox}

\newcommand{\numepisodes}{966}

\newcommand{\myparagraph}[1]{\noindent \textbf{#1}}
\newcommand{\rem}[1]{\texttt{\scriptsize \textcolor{ColComments}{// #1}}}

\hypersetup{
    colorlinks=true,
    citecolor=CornflowerBlue,
    linkcolor=black,
    urlcolor=orange
}

\definecolor{lightgray}{HTML}{EFEFEF}
\definecolor{midgray}{HTML}{C0C0C0}
\definecolor{darkgray}{HTML}{9B9B9B}
\definecolor{coolred}{HTML}{E77475}
\definecolor{coolblue}{HTML}{277C9D}
\definecolor{coolgreen}{HTML}{598938}
\definecolor{coolyellow}{HTML}{FACB77}
\definecolor{coolorange}{HTML}{FF8C00}
\definecolor{coolcyan}{HTML}{3EC3B2}
\definecolor{coollightblue}{HTML}{62BCDD}
\definecolor{coolpurple}{HTML}{976AA3}
\definecolor{ColComments}{HTML}{F7C5A6}

\newcommand{\textreal}[1]{~\colorbox{coolgreen!40}{#1}}

\newcommand{\textmuted}[1]{\scriptsize{\textcolor{darkgray}{#1}}}

\newcommand{\simbox}[1]{\tcbox[on line,colframe=white,boxsep=0pt,left=1pt,right=1pt,top=0pt,bottom=0pt,colback=coolblue!30]{#1}}

\newcommand{\realbox}[1]{\tcbox[on line,colframe=white,boxsep=0pt,left=1pt,right=1pt,top=0pt,bottom=0pt,colback=coolgreen!30]{#1}}

\newcommand{\suppmat}{supp.mat.}

\newcommand{\myrule}{\specialrule{1pt}{0pt}{0pt}}
\newcolumntype{M}[1]{>{\centering\arraybackslash}m{#1}}
\newcolumntype{Y}{>{\centering\arraybackslash}p}
\newcolumntype{Z}{>{\centering\arraybackslash}X}

\newcolumntype{H}{>{\centering\arraybackslash\columncolor{coolblue!10}}X}  %
\newcolumntype{N}{>{\centering\arraybackslash\columncolor{orange!20}}X}
\newcolumntype{R}{>{\centering\arraybackslash\columncolor{coolgreen!10}}X}  %

\def\RR{{\mathbb{R}}}

\def\bF{{\bm{F}}}

\def\bW{{\bm{W}}}

\def\bmf{{\bm{f}}}

\def\bmq{{\bm{q}}}

\begin{document}

\newcommand{\savedtitle}{A scalar per patch from pre-trained ViTs enables fast moving navigation in the real world}
\title{\savedtitle}

\author{Steeven Janny \and
Leonid Antsfeld \and
Christian Wolf}

\authorrunning{Janny et al.}

\institute{NaverLabs Europe, Grenoble, France \\
\email{{name}.{surname}@naverlabs.com}}

\maketitle

\begin{abstract}    
Trained policies for real-world robotics rely on computer vision components, typically in the form of pre-trained visual encoders. These encoders are an essential component and it has been shown that their power does not emerge from training on robotics downstream losses alone. Pre-training with auxiliary losses in the form of computer-vision pre-text tasks is a defining factor and heavily conditions agent performance in robotics tasks. In this unprecedented large-scale study, we ran \numepisodes{} navigation episodes of static point goal navigation in a real-world building for 24km and asked which components really matter for the computer vision aspects of robotics: we evaluate state-of-the art visual encoders in realistic conditions. We explore the usefulness of heterogeneous multi-teacher distillation leading to encoders with multiple different and complementary skills. We investigate how much information from these encoders is necessary for robotics by bottlenecking them in a principled and ``spatially useful'' way and we show that this leads to the emergence of interpretable features linked to affordances. We also argue that training policies on RGB data alone does not lead to an optimal usage of visual features and show this by finetuning policies pre-trained on privileged information. All in all, we paint a more complete picture of what aspects of computer vision are relevant for real-world navigation.
  \keywords{Visual encoders, robotics, navigation}
\end{abstract}

\section{Introduction}
The quest for general-purpose physical intelligence has historically been partitioned into two distinct silos: computer vision, tasked with ``knowing what is where'' and robotics, tasked with mapping those perceptions to motor commands. Computer vision researchers focused on intermediate representations and tasks, such as 3D reconstruction, semantic segmentation, and object detection or tracking, while the robotics community designed controllers to feed on the corresponding representations and features. 

Recent discussions in the field\footnote{See, for instance, {\small \url{https://www.vincentsitzmann.com/blog/bitter_lesson_of_cv}}} suggest that computer vision, as an independent discipline focused on intermediate mappings, may be reaching its "\textit{Bitter Lesson}" moment\footnote{\url{http://www.incompleteideas.net/IncIdeas/BitterLesson.html}}.
Some emerging consensus suggests that the future of perception lies in end-to-end perception-action loops, where the utility of a visual representation is measured solely by its contribution to ``intelligent action''.

\begin{figure}[t] \centering
    \begin{tikzpicture}
        \draw (0, 0) node[anchor=west, inner sep=0] {\includegraphics[width=\textwidth]{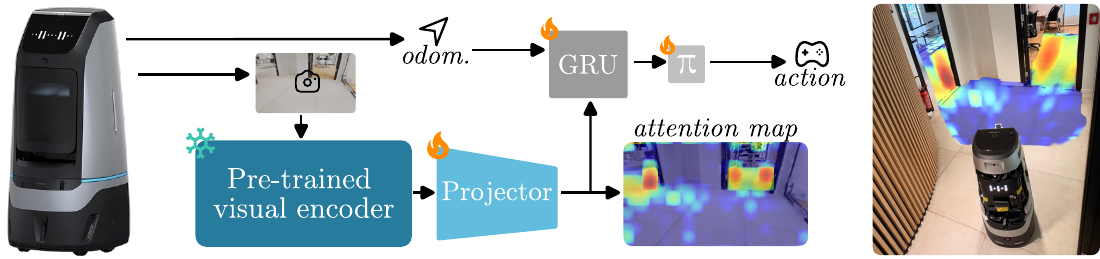}};
        \draw (0.755\textwidth, -1) node[anchor=center, inner sep=0] {(a)};
        \draw (0.77\textwidth, 1.2) node[anchor=center, inner sep=0] {(b)};
    \end{tikzpicture}
    \caption{\label{fig:teaser}We provide an analysis of pre-trained vision encoders for robotic perception in a large-scale study of \numepisodes{} navigation episodes of fast-moving agents in a real environment. (a) we show that interpretable spatial maps emerge through a new proposed bottlenecking layer. These maps seem to encode ``affordances'' and navigation performance is maintained even if they are the only features available to the agent; (b) Integration of multiple maps into a bird's-eye-view.}
    \vspace{-5mm}
\end{figure}

While we agree with this general but probably very long-term trend, current work on robotics heavily builds on the foundations of computer vision. Even the most generalist agents and VLAs \cite{octo2024,pi05,smolVLA2025} are not trained from scratch on downstream robotics data, but rely on pretrained vision encoders whose features are injected into the final model in several training steps. Given the sparsity of robotics data and the difficulty of obtaining it, it is unclear when this will change. In the meantime robotics strongly relies on computer vision in the form of pre-text tasks, pre-training data, and as we will show in this paper, the combination of multiple vision experts.

In this work we provide a large-scale study consisting of an unprecedented \numepisodes{} indoor navigation episodes (and many more performed as preparation) with a real robot in realistic conditions. By that we mean  fast-moving agents navigating at \SI{0.7}{\meter\per\second} and which can be deployed to solve real-life tasks. The agents only use RGB input for navigating and do not have access to Lidar or depth cams. Moving at this speed requires robust perception and decision making. The goal of this work is to study the role of visual perception in navigation from several angles and we ask the following questions:
\begin{description}[nosep,itemsep=1mm,labelindent=0mm,leftmargin=7mm,topsep=0mm]
    \item[(1)] Can perception be learned from actions alone? We show that direct training from scratch of visual encoders leads to bad performance, ie. difficult transfer to real physical robots, as the learning algorithm overfits the learned representations to visual artifacts and spurious correlations. We  argue that the usage of pre-trained vision encoders combined with realistic motion modelling~\cite{bono2024learning,janny2025} has closed the sim2real gap to negligible size.

    \item[(2)]  What kind of pre-training is most suitable for navigation downstream tasks? We compare several different pre-training strategies including MAE \cite{mae2022,vc12024} and the hybrid combination of self-distillation \cite{dino2021} and iBOT \cite{ibot2022} used in \textit{DinoV2} \cite{dinov22024} and \textit{DinoV3} \cite{dinov32025}. 

    \item[(3)] Can robotics / navigation tasks benefit from complementary and heterogenous capabilities stemming from training on different high-level tasks? We explore perception with highly versatile encoders like \textit{AM-Radio} \cite{radio25} and \textit{Dune} \cite{dune2025} distilled from multiple sources including 3D reconstruction with \textit{MASt3R} \cite{mast3r}, human mesh recovery with \textit{MHMR} \cite{mhmr2024} and general vision features from \textit{DinoV2} \cite{dinov22024}.

    \item[(4)] How much information encoded in these visual features is necessary for the downstream task? We show that the information flow can be restricted to only 1 scalar per image patch. Furthermore, by shaping the information flow such that it corresponds to spatial attention distributions, interpretable features and affordances emerge without loss of navigation quality.

    \item[(5)] Is privileged sensor information helpful during training? We show that pre-training agents with RL on Lidar-like input with a full 360° view can provide gains when finetuning and deploying them with input from forward facing RGB only. 
\end{description}

\vspace{1mm}
\noindent 
In summary, this study offers a comprehensive analysis of how various visual pre-training strategies dictate the navigational performance and reliability of fast-moving robots in complex settings.

\section{Related works}

\myparagraph{Visual encoders in robotics} originate from vision transformers trained with masked image modelling \cite{mae2022}. This approach can be adapted to robotic settings by training on large collection of ego-centric images gathered across platforms (\textit{VC1} \cite{vc12024}, \textit{MVP} \cite{qian2025mvp}). Self-distilled and iBOT-based models \cite{ibot2022}, such as the \textit{Dino} series \cite{dino2021, dinov22024,dinov32025} are widely adopted due to their strong transfer performance and availability in multiple scales compatible with on-board computation. In particular, \textit{DinoV2} has become a common backbone for downstream robotic tasks \cite{garg2024robohop,wang2024dfields,zeng2024poliformer}, motivating world models that operate in its feature space, both at the input and output levels (e.g. \textit{Dino-WM} \cite{zhou2025dinowm} and \textit{Dino Foresight} \cite{karypidis2026dinoforesight}). Complementary to general-purpose encoders, models tailored to 3D geometry (e.g. \textit{DUSt3R} \cite{wang2024dust3r}, \textit{MASt3R} \cite{mast3r}, \textit{VGGT} \cite{vggt2025}) provide spatially structured features that benefit navigation and manipulation tasks \cite{liu2025citywalker, CrocoNav2024}. Finally, multiple specialized encoders can be distilled into a single ViT to produce compact representations aggregating diverse capabilities (e.g., \textit{AM-Radio}~\cite{radio25}, \textit{Dune}~\cite{dune2025}).

\myparagraph{Visual navigation} has been classically solved in robotics using explicit modelling~\cite{burgard1998interactive,macenski2020marathon,marder2010office}, which requires solutions for mapping and localization~\cite{bresson2017simultaneous, labbe19rtabmap,thrun2005probabilistic}, 
for planning~\cite{konolige2000gradient, sethian1996fast} and for low-level control \cite{fox1997dynamic,rosmann2015timed}. These methods depend on accurate sensor models, filtering, dynamical models and optimization. End-to-end trained models directly map input to actions and are typically trained with RL~\cite{DBLP:conf/iclr/JaderbergMCSLSK17,mirowski17learning,zeng2024poliformer,uppal2024spin} or IL~\cite{DBLP:conf/nips/DingFAP19}. They learn flat representations~\cite{bono2024learning}, occupancy maps~\cite{Chaplot2020Learning}, semantic maps~\cite{chaplot2020object,monaci2024zerobev}, latent metric maps~\cite{DBLP:conf/pkdd/BeechingD0020,Henriques_2018_CVPR,DBLP:conf/iclr/ParisottoS18}, topological maps~\cite{BeechingECCV2020,Chaplot_2020_CVPR,shah2022viking}, explicit episodic memory~\cite{chen_think_2022,du2021vtnet,Fang_2019_CVPR,reed_generalist_2022}, implicit representations~\cite{Marza2022NERF} or navigability~\cite{Mole2024}.
Our study targets recurrent policies trained end-to-end additional motion model in simulators equipped with realistic motion.

\myparagraph{Querying representations in computer vision and embodied AI}
Feature maps produced by CNN- or ViT-based encoders are commonly extracted through flattening (aggregation of patch/cell embeddings), global average pooling, ROI-pooling for object-centric tasks (introduced in \textit{Faster-R-CNN}~\cite{ren2016fasterrcnn}), or dedicated class tokens that can be generalized to attention-based read-outs (e.g., \textit{DETR}~\cite{carion2020detr}). Multi-scale features are combined in \textit{Feature Pyramid Networks}~\cite{lin2017fpn} and \textit{Hypercolumns}~\cite{hariharan2015hypercolumns}, while place recognition relies on \textit{NetVLAD / VLAD}-like aggregation~\cite{netvlad2016}, where local descriptors are assigned to learned centroids and residuals accumulated. More closely related to our approach, several works query frozen visual representations with cross-attention and learned embeddings, such as \textit{Slot Attention}~\cite{slotattention2020} and \textit{QFormer}~\cite{qformer2023}. In embodied AI, end-to-end policies typically reuse similar pooling strategies before passing features to recurrent or transformer-based policies. Goal-conditioned tasks (e.g., image-goal navigation) rely on temporal stacking, channel-wise stacking, or cross-attention~\cite{shah2023vint,sun2024fgprompt,CrocoNav2024,monaci2026doesreallymatterimage}. Vision-language-action models project visual features into the token space of VLMs or LLMs by representing images as sets of CNN or ViT tokens (e.g., \textit{Open-VLA}~\cite{openvla2024}, \textit{SmolVLA}~\cite{smolVLA2025}, \textit{Octo}~\cite{octo2024}, $\pi_0$~\cite{pi0}), which are then integrated by attention layers operating over single or multiple time steps. Prior work \cite{sax2020learning} evaluates mid-level priors solely in simulation without real-world validation, and \cite{ai2023invariance} study sim-to-real representation transfer but rely on handcrafted features like depth and navigability. In contrast, we use general-purpose pretrained ViTs and compress their features using learned projectors without task-specific engineering.

\section{Experimental Setup}
\label{sec:experiments}

\setlength{\intextsep}{0pt}%
\setlength{\columnsep}{8pt}%
\begin{wrapfigure}{l}{5.5cm} 
    \vspace{-3mm}
    \centering
    \includegraphics[width=\linewidth]{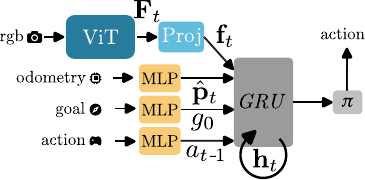}
    \caption{\label{fig:agent}The navigation agent used in the experimental setup uses recurrent memory. It is based on \cite{janny2025} but uses RGB only input to collect information on the scene structure.}     
\end{wrapfigure} 
For this study we build on the agent from Janny et al. \cite{janny2025}, which has been optimized for real-world scenarios and fast-moving agents trained end-to-end in simulation, with a primary focus on minimizing the sim-to-real gap. To make the paper self-contained, we will  describe the agent in this section. We choose static point-goal navigation as a standard task for \emph{visual navigation}: It lets us isolate the impact of vision on policy learning without additional complexity from language grounding or image-goal matching. Although broader tasks require different reasoning mechanisms, static point-goal captures core navigation skills such as obstacle perception, scene understanding, short-horizon memory, and path execution.

At each discrete time step $t$, the agent receives  sensor readings, which include an RGB image $\mathbf{I}_t$. In the original work \cite{janny2025,bono2024learning}, the agent also received a Lidar-like ``\textit{scan}'' vector which represents depth ranges analogous to Lidar data, but which was actually extracted from four directional \textit{RealSense} depth sensors. In our work, we do not use this sensor information directly related to the 3D structure of the environment and \textbf{show that the agent can navigate from RGB being the only input alone}, without Lidar or depth.

The task is \textit{Static PointGoal} in real environments, i.e., the agent's objective is to reach a target defined by polar coordinates relative to its  pose at episode start. In other words, the goal vector is \textit{not} updated when the agent moves and stays constant, for this reason we denote it as $\mathbf{g}_0$. The agent operates without an external map, relying instead on onboard ego-motion estimation. To facilitate goal-tracking, we provide a localization signal relative to the starting position through odometry $\hat{\mathbf{p}}_t$ derived from wheel encoders and IMUs.
This signal includes position and velocity estimates. 

The agent runs at 3Hz and outputs actions from a discretized action space consisting of 28 velocity pairs $(a_v, a_\omega)$. The linear velocity $a_v$ is sampled from $\{0, 0.2, 0.4, 0.7\}$ \unit{\meter\per\second}, while angular velocity $a_\omega$ ranges from $\{-3, \dots, 3\}$ \unit{\radian\per\second} in integer increments. To handle the partial observability of the environment, the agent maintains a recurrent hidden state $\mathbf{h}_t$:
\setlength{\abovedisplayskip}{5pt}
\setlength{\belowdisplayskip}{5pt}
\begin{equation}
\begin{array}{lll}
\arraycolsep=1.4pt
\mathbf{F}_t & 
= v(\mathbf{I}_t) &
\rem{Perception / Vision}
\\
\mathbf{h}_t & 
= d(\mathbf{h}_{t-1}, \mathbf{F}_t, \mathbf{g}_0, \hat{\mathbf{p}}_t, \mathbf{a}_{t-1}) &
\rem{State update}
\\
\mathbf{a}_t & = 
\pi(\mathbf{h}_t) &
\rem{Policy}
\\
\hat{\mathbf{g}}_t & 
= l(\mathbf{h}_{t}) &
\rem{Auxiliary loss f. goal prediction}
\end{array}
\end{equation}

\noindent
In this model, the visual input is processed by a pre-trained ViT $v$ extracting features $\mathbf{F}_t$ from the input RGB image $\mathbf{I}_t$, which will be main object of study in this work. Memory is updated by $d$, a two-layer (GRU) \cite{cho-etal-2014-learning}, previous actions are processed with an embedding layer and other state inputs are mapped through MLPs. The final policy $\pi$ is implemented as a linear head. As in \cite{janny2025}, since the goal $\mathbf{g}_0$ is fixed in the initial frame, we assist the agent’s spatial reasoning by adding an auxiliary linear head $l$ that predicts the instantaneous relative goal $\hat{\mathbf{g}}_t = l(\mathbf{h}_{t})$, supervised during training via simulation ground truth.

The model is trained with RL, in particular PPO~\cite{schulman2017proximal}, and with a reward inspired by \cite{chattopadhyay2021robustnav,janny2025},
$
r_t=\mathrm{R} \cdot \mathbb{I}_{\text {success}} -\Delta_t^{\mathrm{Geo}} -\lambda -\mathrm{C} \cdot \mathbb{I}_\text{collision}
$, where $R{=}2.5$, $\Delta_t^{\mathrm{Geo}}$ is the gain in geodesic distance to the goal, a slack cost $\lambda{=}0.01$ encourages efficiency, and a cost $C=0.1$ penalizes each collision without ending the episode.

\vspace{2mm}
\myparagraph{Modelling Realistic Dynamics in Real Environments}.
We corroborate the importance of modelling realistic motion dynamics in simulation introduced in \cite{bono2024learning,janny2025}. We replace the standard ``teleportation'' motion in the Habitat simulator \cite{Savva_2019_ICCV} with a second-order dynamical motion model, whose parameters were identified from real-world trajectories with our robot. This ensures that simulation reflects authentic inertial effects: the simulated robot accelerates, brakes and turns as the physical one would in the same conditions. \textbf{Realistic motion modelling is crucial}: since the robot moves at \SI{0.7}{\meter\per\second} and takes decisions at \SI{3}{\hertz}, it needs to be able to correctly anticipate its future trajectory, as any potential corrections for an action can be done after \SI{333}{\milli\second} at the earliest.

\subsection{Experimental Protocol and Metrics}

We train the agent with the Habitat simulator \cite{Savva_2019_ICCV}, modified for realistic motion as described above, and on the \textit{HM3D} \cite{ramakrishnan2021hm3d} training set of 800 scenes 3D-scanned by a Matterport scanner. We evaluate our models across two primary domains to balance statistical significance with real-world validity:
\begin{description}
    \item[\realbox{Physical Deployment (``Real'')}]: Tests are conducted on a ``\textit{Rookie}'' robot (cf. Fig. \ref{fig:teaser}) within a designated office environment. We report results across 14  episodes per experiment and we systematically report avg$\pm$std-dev over 3 runs, ie. 42 real navigation episodes per evaluated configuration.
    \item[\simbox{Augmented Simulation (``Sim'')}]: We utilize the simulator integrated with our identified dynamics model, which makes evaluations highly predictive of behavior in the real environment. Evaluations are performed on the HM3D \cite{ramakrishnan2021hm3d} validation set (2,500 episodes).
\end{description}

\noindent
All tables have color-coded backgrounds indicating the evaluation setting.
\vspace{1mm}

\myparagraph{Performance metrics.}
Success is defined by the agent reaching within \SI{0.2}{\meter} of the goal in simulation (with zero commanded motion) or \SI{1}{\meter} in real-world trials. We report the usual metrics for navigation~\cite{DBLP:journals/corr/abs-1807-06757}: (1) \textit{Success Rate} (SR), ie. the ratio of successfully completed episodes (2) spatial efficiency as \textit{Success weighted by Path Length} (SPL), (3) temporal efficiency as 
\textit{Success weighted by Completion Time} (SCT) \cite{yokoyama2021success}:
\setlength{\abovedisplayskip}{5pt}
\setlength{\belowdisplayskip}{5pt}
\begin{equation}
\textit{SR} = \frac{1}{N} \sum_{i=1}^N \mathbb{I}_{i},
\ 
\textit{SPL} = \frac{1}{N} \sum_{i=1}^N \mathbb{I}_{i} \frac{\ell_i^*}{\max(\ell_i, \ell_i^*)}, 
\ 
\textit{SCT} = \frac{1}{N} \sum_{i=1}^N  \mathbb{I}_{i} \frac{c_i^*}{\max(c_i, c_i^*)}
\end{equation}
where $\mathbb{I}_{i}$ is a success indicator for the $i^{th}$ episode, $\ell_i$ and $\ell_i^*$ are the navigated and shortest path, respectively, and $c_i$ and $c_i^*$ are the elapsed time and the theoretically optimal time, respectively. Due to real-world hardware and sensor limitations, we use different success thresholds: simulation uses ground-truth positioning with a \SI{0.2}{\meter} threshold and real-world evaluation relies on noisy onboard sensors and AMCL with a \SI{1.0}{\meter} threshold.

\section{Experimental Results}
\label{sec:setup}
\myparagraph{Lineup of vision models}. We tested perception on several widely used vision encoders and state-of-the-art models, if not stated otherwise as ViT/Base:
\begin{description}
    \item[DinoV2 \cite{dinov22024}] is trained on large procedurally curated dataset of 142M images with several self-supervised losses, including self-distillation from Dino \cite{dino2021} and the patch-wise iBOT \cite{ibot2022}. It provides general features that are highly transferable across diverse downstream tasks without fine-tuning.
    \item[DinoV3 \cite{dinov32025}] scales its predecessor up to larger model architectures and even more massive datasets, incorporating long-sequence training and improved stability during the distillation process.
    \item[VC1 \cite{vc12024}] uses MAE \cite{mae2022} for training. It specifically targets robotics in its datasets combining 4.3M images stemming from ImageNet and 4000h of egocentric video from 7 sources. 
    \item[AM-RADIO v2.5\cite{radio25}] is distilled from heterogeneous sources such as \textit{CLIP} \cite{clip2021}, \textit{DinoV2} \cite{dinov22024}, and \textit{SAM} \cite{sam2023}, aiming to combine the best of semantic, geometric, and segment-level features into a single efficient backbone.
    \item[DUNE \cite{dune2025}] is also a distilled model but focuses on unifying representations across diverse environments by distilling from multiple specialized teacher models into a compact, robust architecture. It distills from \textit{DinoV2} \cite{dinov22024}, \textit{MASt3R} \cite{mast3r} and \textit{MHMR} \cite{mhmr2024} and beats \textit{MASt3R} on its own task, map-free localization, which we argue is particularly relevant for robotics tasks.
    \item[ResNet from scratch \cite{resnet2016}] is a non-pretrained baseline, a ResNet-18  initialized from scratch and entirely trained with the downstream RL loss.
\end{description}

\begin{figure}[t] \centering
    \includegraphics[width=0.9\textwidth]{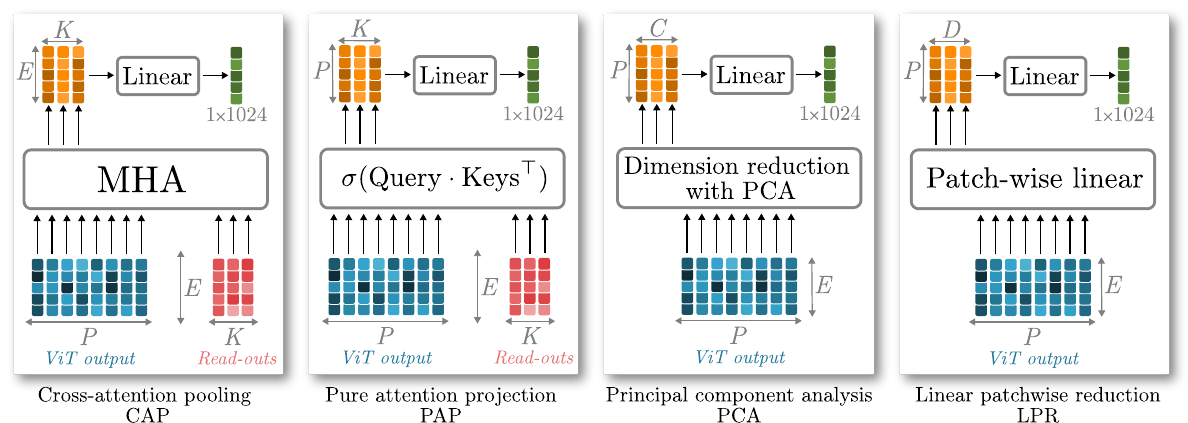}
    \caption{\label{fig:projectors_scheme}\textbf{Architecture of projection heads} -- inserted between the dense, high-dimensional output of the pretrained visual encoders, and the recurrent memory of the agent. The LPR and PAP maintain spatial information by computing patch-wise embeddings. The PCA projects each patch on the $C$ principal components, with no learnable parameter apart from the final linear layer.}    
\end{figure}

\myparagraph{Lineup of projection heads.} Apart from the ResNet from scratch, all vision encoders have been frozen during RL training. They produce a feature tensor $\mathbf{F}_t \in \mathbb{R}^{P{\times}E}$ where $P$ is the number of patches and $E$ the embedding dimension. When useful we will denote individual patch features for patch $p$ by $\mathbf{F}_{t,p}$. We project / pool the features with a trainable layer into a lower dimensional vector $\mathbf{f}_t$ and explore several projection mechanisms (also illustrated in Fig. \ref{fig:projectors_scheme}): 
\begin{description}
    \item[Cross-attention pooling] (CAP) is a widely used standard features extraction method in computer vision, a.k.a ``\textit{Perceiver-Resampler}'' \cite{perceiver,flamingo} in the context of VLMs. It employs a set of $K$ learned query embeddings $\mathbf{q}_k$ to cross-attend into the feature tokens with multi-head attention,
    $
    \mathbf{f}_t = \{\text{MHA}(\bmq_k, \bF_t)\}_{k=1..K}.
    $
    \item[Pure Attention Projection:] (PAP) we introduce a variant of attention pooling which entirely discards the value projections and where the attention distribution has been ``\textit{elevated}'' from an internal means of computation to the actual representation passed to the downstream agent. For a single head with a query projection matrix $\mathbf{W}_{\text{query}}$ and key projection matrix $\mathbf{W}_{\text{key}}$, this can be written as
    $
    \label{eq:hermes_layer}
    \mathbf{f}_t = \{\sigma(\mathbf{q}_k \mathbf{W}_{\text{query}} \cdot(\mathbf{F}_t \mathbf{W}_{\text{key}})^{\top})\}_{k=1...K},
    $
    where $\sigma$ is the softmax function. As we will see in the experiments further below, this will keep the downstream performance of full attention pooling while at the time lead to arguably quite interpretable vision features.
    \item[Principal Component Analysis] (PCA) projects each patch on C principal components, with no learnable parameter apart from the final linear layer.
    \item[Linear Patchwise Reduction] (LPR) projects each patch feature vector in $\mathbf{F}_t$ down to a low dimensional vector of size $D$, $\mathbf{f}_{t,p} = \mathbf{W} \mathbf{F}_{t,p}$ where $\mathbf{W}$ is a weight matrix.
\end{description}

\begin{table}[t] \centering
\setlength{\aboverulesep}{0pt}
\setlength{\belowrulesep}{0pt}
    \footnotesize
     \begin{tabularx}{\textwidth}{l| HHH || RRR }
        \myrule
        \multicolumn{7}{c}{\textit{Trained from scratch (500M steps)}} \\
        \myrule
        \cellcolor{darkgray}ViT & \multicolumn{3}{c||}{\cellcolor{coolblue!40}Sim (HM3D/2.5k)} & \multicolumn{3}{c}{\cellcolor{coolgreen!40}Real (TestBuilding/14) } \\ 
        \rowcolor{midgray}\cellcolor{darkgray}Backbone & \cellcolor{coolblue!30}SCT & \cellcolor{coolblue!30}SPL & \cellcolor{coolblue!30}SR & \cellcolor{coolgreen!30}SCT & \cellcolor{coolgreen!30}SPL & \cellcolor{coolgreen!30}SR \\ \myrule
        \cellcolor{lightgray}DinoV2  & 16.1 & 49.8 & 63.4 & 11.8 \textmuted{$\pm 1.3$} & 37.2 \textmuted{$\pm 3.9$} & 64.3 \textmuted{$\pm 5.8$} \\
        \cellcolor{lightgray}VC-1     & 28.9 & 72.2 & 86.3 & 24.3 \textmuted{$\pm 1.9$} & \textbf{63.4} \textmuted{$\pm 4.8$} & 95.2 \textmuted{$\pm 6.7$}  \\
        \cellcolor{lightgray}Dune     & \textbf{33.4} & \textbf{76.9} & \textbf{90.7}& 26.8 \textmuted{$\pm 2.0$} & 62.5 \textmuted{$\pm 3.4$} & 95.3 \textmuted{$\pm 3.3$} \\  
        \cellcolor{lightgray}AM-Radio & 27.6 & 68.2 & 85.2 & \textbf{29.3} \textmuted{$\pm 0.7$} & \textbf{63.4} \textmuted{$\pm 0.5$} & \textbf{100} \textmuted{$\pm 0.0$} \\  
        \cellcolor{lightgray}DinoV3   
            & \multicolumn{3}{c||}{\cellcolor{coolblue!10}{0.0 $\rightarrow$ \textit{see Table \ref{tab:dinov3}}}}  
            &\multicolumn{3}{c}{\cellcolor{coolgreen!10}{\textit{not evaluated}}}  \\ 
        \cellcolor{lightgray}\textit{*ResNet}              & 15.0 & 41.0 & 53.5 & \multicolumn{3}{c}{\cellcolor{coolgreen!10}{\textit{not evaluated}}}   \\ \myrule
        \multicolumn{7}{r}{\tiny{*Half-width ResNet trained from scratch}}
    \end{tabularx}
    \caption{\label{tab:from_scratch}\textbf{Comparison of vision encoders} -- kept frozen while the remaining agent is RL-trained from scratch for 500M steps. Projection heads are Cross-attention pooling (CAP). Real experiments: 3 runs $\times$14 episodes per model. We report ``avg \textcolor{darkgray}{$\pm$ std}'' \realbox{(168 real nav episodes in this table)}.}
    \vspace{-8mm}
\end{table}

\myparagraph{How to train your Encoder.}
Table \ref{tab:from_scratch} shows navigation results in \simbox{simulation} with realistic motion (left) and in the \realbox{real environment} (right) and   for the lineup of different vision encoders. The ViTs were kept frozen, but the rest of the agent (i.e. ViT projection head, modality encoders, recurrent memory and policy) were trained from scratch for 500M steps. We can see that the vision encoder baseline trained from scratch with RL did not perform at all (SR${=}53.5\%$), which we link to the combination of the weak learning signal provided by RL and the difficulty of training an agent with realistic motion in the augmented simulator\footnote{Training an agent with classical ``teleportation'' actions, eg. the action space ={\small \tt \{move forward 0.25m, turn left $10^{\circ}$, turn right $10^{\circ}$, stop\}}, leads to relatively high training success rates but also a high sim2real gap and low transferability.}.

Using pre-trained ViTs with simple Cross-attention pooling (CAP) leads to quite high performance, and we can see that the performance metrics are comparable between the 2,500 episodes evaluated in simulation and the 3${\times}14$ episodes evaluated with the real robot in the real environment. This suggests our main finding: we argue that \textbf{the usage of pre-trained vision encoders combined with realistic motion modelling \cite{bono2024learning,janny2025} has closed the sim2real gap to neglectable size}. Navigation of fast-moving agents can indeed be trained purely in simulation without any real data involved.

The choice of vision encoder does impact performance and generalist ViTs such as DinoV2 and DinoV3 under-perform. We will investigate the failure of DinoV3 at the end of the paper. The two encoders using distillation from heterogeneous teachers perform well, in particular in the real experiments. We conjecture that mixing multiple tasks can overcome shortcomings of pure self-supervised learning. The performance of VC-1 was surprising to us, as the MAE loss is not known to outperform self-distillation. We suspect that the good performance might stem from the data mix of 4.3M images specially targeting robotics.

\begin{table}[t]
    \centering
    \setlength{\tabcolsep}{0.5em}
    \begin{tabularx}{\textwidth}{l| HHH || RRR }
        \myrule
        \multicolumn{7}{c}{\textit{Finetune from an agent receiving a Lidar-like input (+100M steps)}} \\
        \myrule
        \cellcolor{darkgray}ViT & \multicolumn{3}{c||}{\cellcolor{coolblue!40}Sim} & \multicolumn{3}{c}{\cellcolor{coolgreen!40}Real} \\ 
        \rowcolor{midgray}\cellcolor{darkgray}Backbone & \cellcolor{coolblue!30}SCT & \cellcolor{coolblue!30}SPL & \cellcolor{coolblue!30}SR & \cellcolor{coolgreen!30}SCT & \cellcolor{coolgreen!30}SPL & \cellcolor{coolgreen!30}SR \\ \myrule
        
        \cellcolor{lightgray}DinoV2   & 31.6 & 74.3 & 88.0 & 27.5 \textmuted{$\pm 4.2$} & 57.2 \textmuted{$\pm 9.4$}& 97.6 \textmuted{$\pm 3.3$} \\
        \cellcolor{lightgray}VC-1     & 32.4 & 76.1 & \textbf{90.3} & 28.3 \textmuted{$\pm 2.4$} & \textbf{65.0} \textmuted{$\pm 4.4$} & \textbf{100}  \textmuted{$\pm 0.0$} \\
        \cellcolor{lightgray}Dune     & \textbf{34.1} & \textbf{77.5} & 89.8 & \textbf{30.1}  \textmuted{$\pm 1.6$} & 64.1  \textmuted{$\pm 1.7$} & \textbf{100}  \textmuted{$\pm 0.0$} \\
        \cellcolor{lightgray}AM-Radio & 30.4 & 72.0 & 87.4 & 25.6 \textmuted{$\pm 0.7$} & 56.0  \textmuted{$\pm 1.5$} & 92.9  \textmuted{$\pm 5.8$} \\
        \cellcolor{lightgray}DinoV3   & \multicolumn{3}{c||}{\cellcolor{coolblue!10}{0.0 $\rightarrow$ \textit{see table \ref{tab:dinov3}}}}   & \multicolumn{3}{c}{\cellcolor{coolgreen!10}{\textit{not evaluated}}}   \\  
        \cellcolor{lightgray}\textit{*ResNet}              & 14.1   & 38.0   & 48.4   &  \multicolumn{3}{c}{\cellcolor{coolgreen!10}{\textit{not evaluated}}}  \\ 
        \cellcolor{lightgray}\textit{Lidar} (n.c.)          & \textit{38.1}   & \textit{81.9}   & \textit{94.6}   & \textit{36.5 \textmuted{$\pm 0.7$}} & \textit{76.5 \textmuted{$\pm 2.7$}} & \textit{100 \textmuted{$\pm 0.0$}} \\ \myrule
        \multicolumn{7}{r}{\tiny{*Half-width ResNet trained from scratch}}
    \end{tabularx}
    \caption{\textbf{Comparison of vision encoders} -- where the agent is initialized from a pre-trained model solving the same navigation task, yet using a 2D 360$^\circ$ Lidar-like scan instead of the RGB input. Models are fine-tuned on RGB input for 100M steps \realbox{(210 real nav episodes in this table)}. For reference, we also report the metrics for the Lidar-like agent we finetuned from, despite not being comparable.}
    \label{tab:from_mibi}
    \vspace{-5mm}
\end{table}

\begin{figure}[t]
    \centering
    \begin{tikzpicture}[font=\footnotesize]
        \draw (0, 0) node[anchor=west, inner sep=0] {\includegraphics[width=\textwidth]{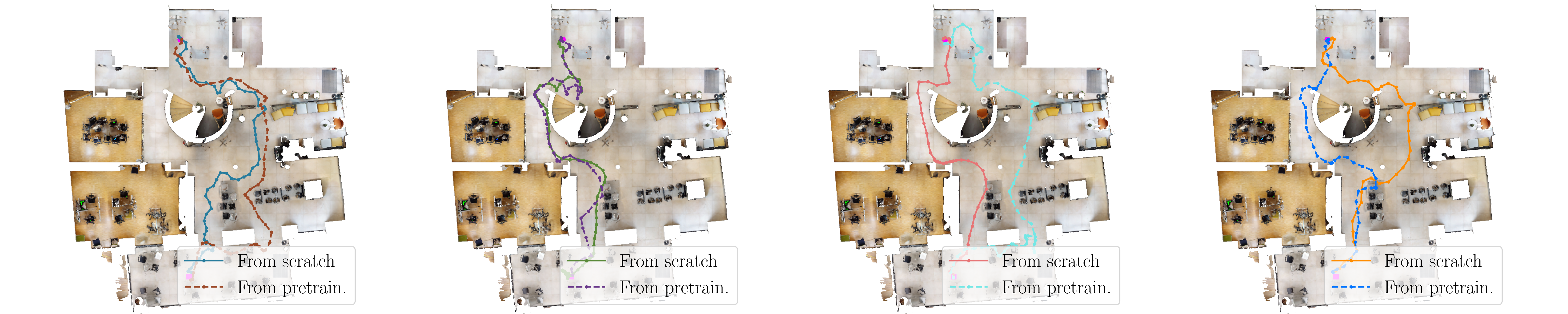}};
        \draw (0.13\textwidth, 1.4) node[anchor=center, inner sep=0] {\textbf{DinoV2}};
        \draw (0.38\textwidth, 1.4) node[anchor=center, inner sep=0] {\textbf{VC1}};
        \draw (0.62\textwidth, 1.4) node[anchor=center, inner sep=0] {\textbf{Dune}};
        \draw (0.86\textwidth, 1.4) node[anchor=center, inner sep=0] {\textbf{AM-Radio}};
    \end{tikzpicture}
    \caption{\textbf{Comparison of \textreal{real trajectories}} -- between agents trained from scratch or finetuned from a pretrained baseline. The agent is tasked to navigate from \textcolor{magenta}{\LARGE{$\bullet$}} to \textcolor{magenta}{$\blacksquare$}.}
    \label{fig:real_traj_versus}
    \vspace{-5mm}
\end{figure}
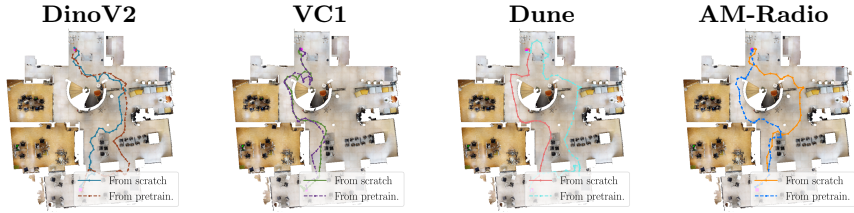

\myparagraph{Pre-train with privileged 3D information.}
Table \ref{tab:from_mibi} shows experiments with agents using the same vision encoders, but this time the rest of the agent is not trained from scratch. Instead, we explore the availability of privileged information available during in simulation. We first train an agent which gets its RGB input replaced by a 2D 360$^\circ$ Lidar-like scan for 850M steps. This input signal is more directly related to the 3D scene structure and allows for easy exploitation by the agent with the RL learning signal. In a second phase, we finetune this agent for 100M steps on vision only input and Cross-attention pooling (CAP).

We observe a boost in metrics by most agents with the exception of the one using the AM-Radio encoder, which translates into better behavior during in the real-world (figure \ref{fig:real_traj_versus}). Information from the privileged agent seems to transfer to the vision-only agents although the actual vision encoders were kept frozen. This experiment also substantially decreased the differences between the vision encoders, which provides evidence that the content of their features is useful for robotics in a similar degree. However, there might be differences in the difficulty of decoding the necessary information by an agent trained with weak RL losses.

\begin{figure}[t]
    \centering
    \includegraphics[width=0.9\linewidth]{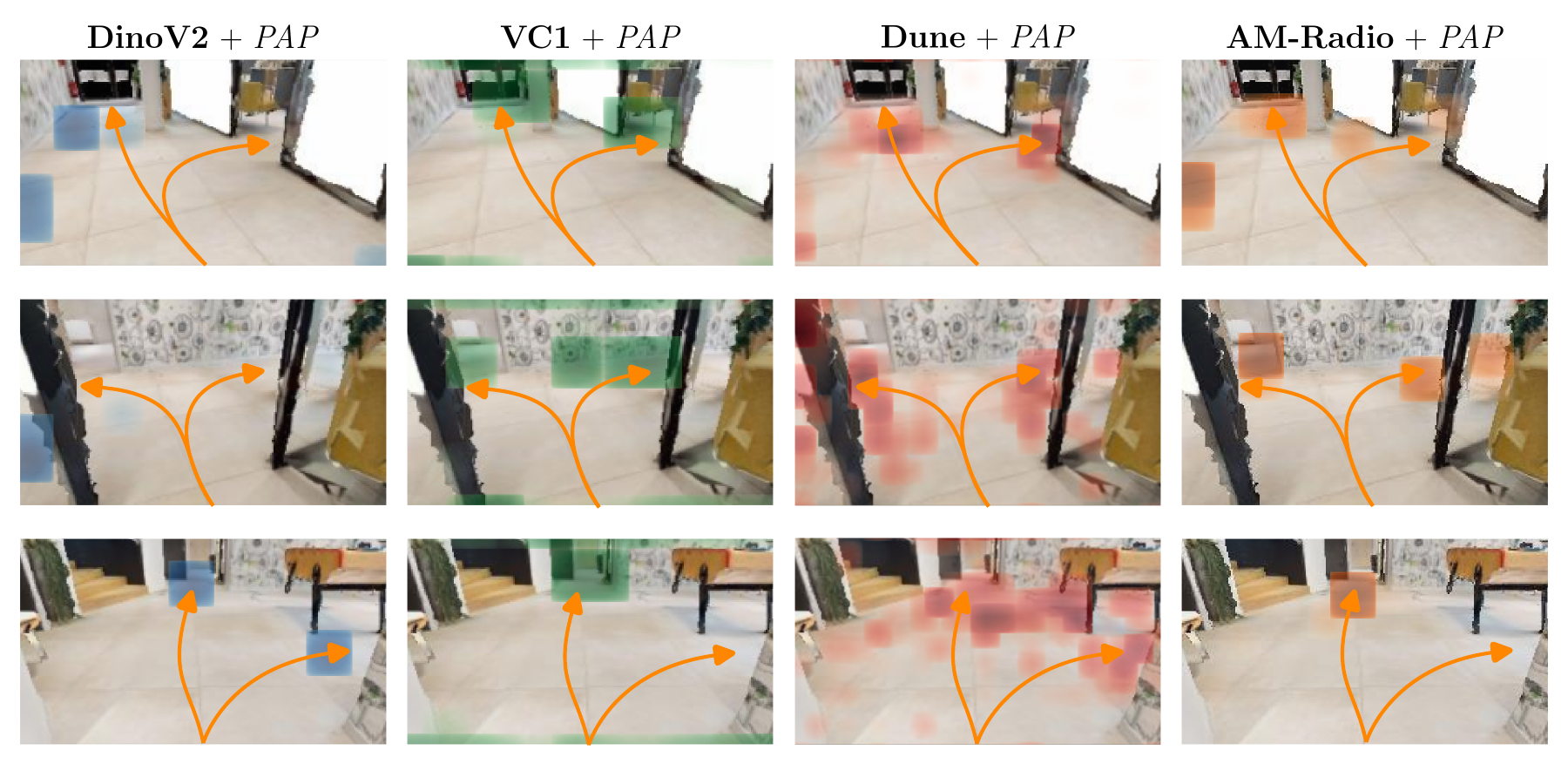}
    \caption{\label{fig:affordances}\textbf{Affordances emerge with Pure Attention Projection}, ie. interpretable maps which are the \textbf{only information on the scene provided to the agent}. The shown map corresponds to one of the read-out tokens ($2^{nd}$ column in Fig. \ref{fig:accummaps}), which we associate with potential directions accessible to the robot (manually labeled with \textcolor{orange}{$\nearrow$}.)}
    \vspace{-5mm}
\end{figure}

\begin{figure}[p]
    \centering    \includegraphics[width=\linewidth]{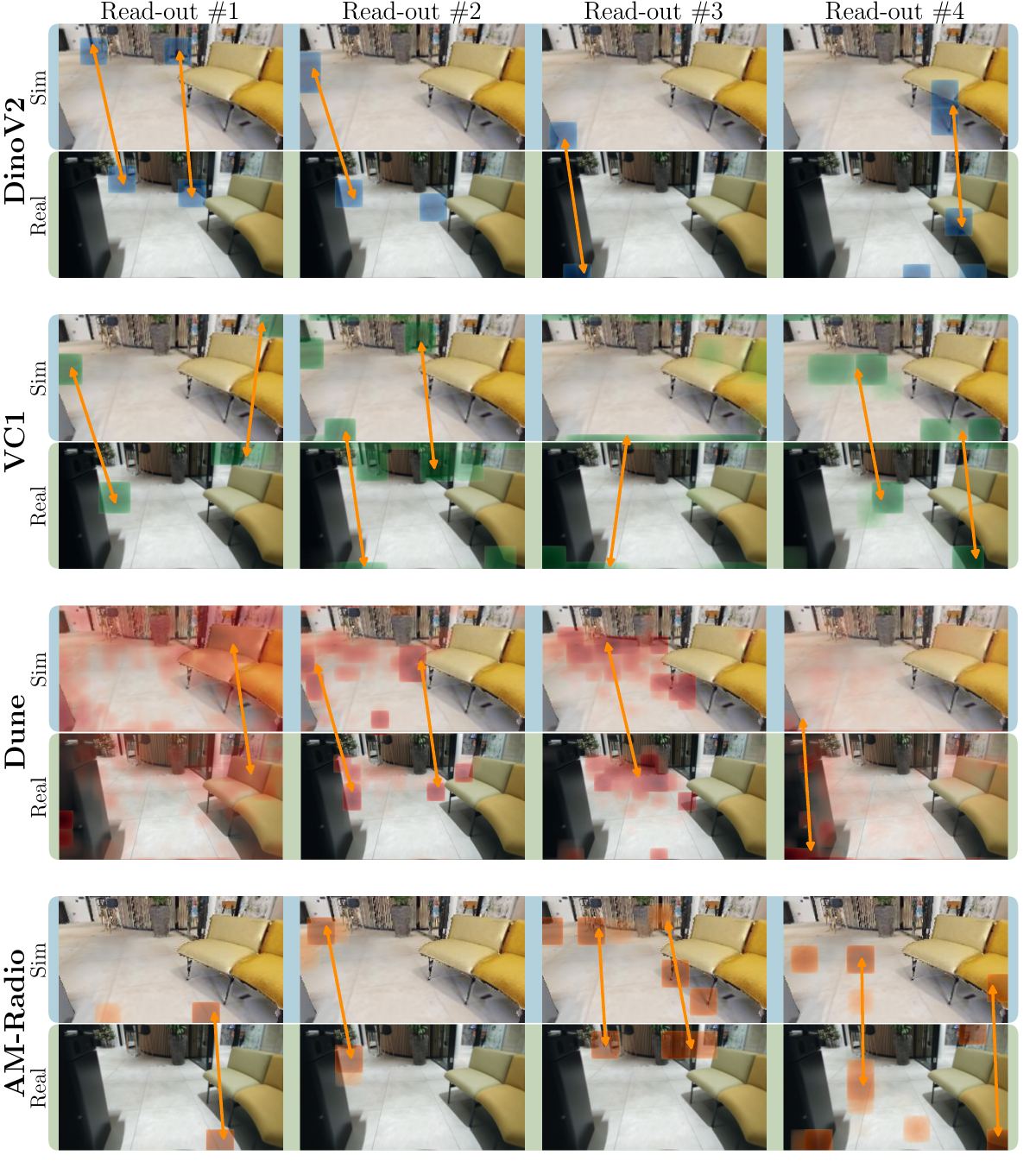}
    \caption{\label{fig:sim2real}\textbf{Low sim2real gap visualized with Pure Attention Projection} 
    for different visual encoders. Each column shows the output given by one read-out token. We aligned viewpoints between real images captured by the robot and rendered views from the simulator, and compare the attention distribution. All ViTs seem to attend to similar locations despite difference in lighting condition and texture. Corresponding regions of attention are manually labeled with \textcolor{orange}{$\nearrow$}. }
    \vspace{-5mm}
\end{figure}

\begin{figure}[t]
    \centering
    \includegraphics[width=0.9\linewidth]{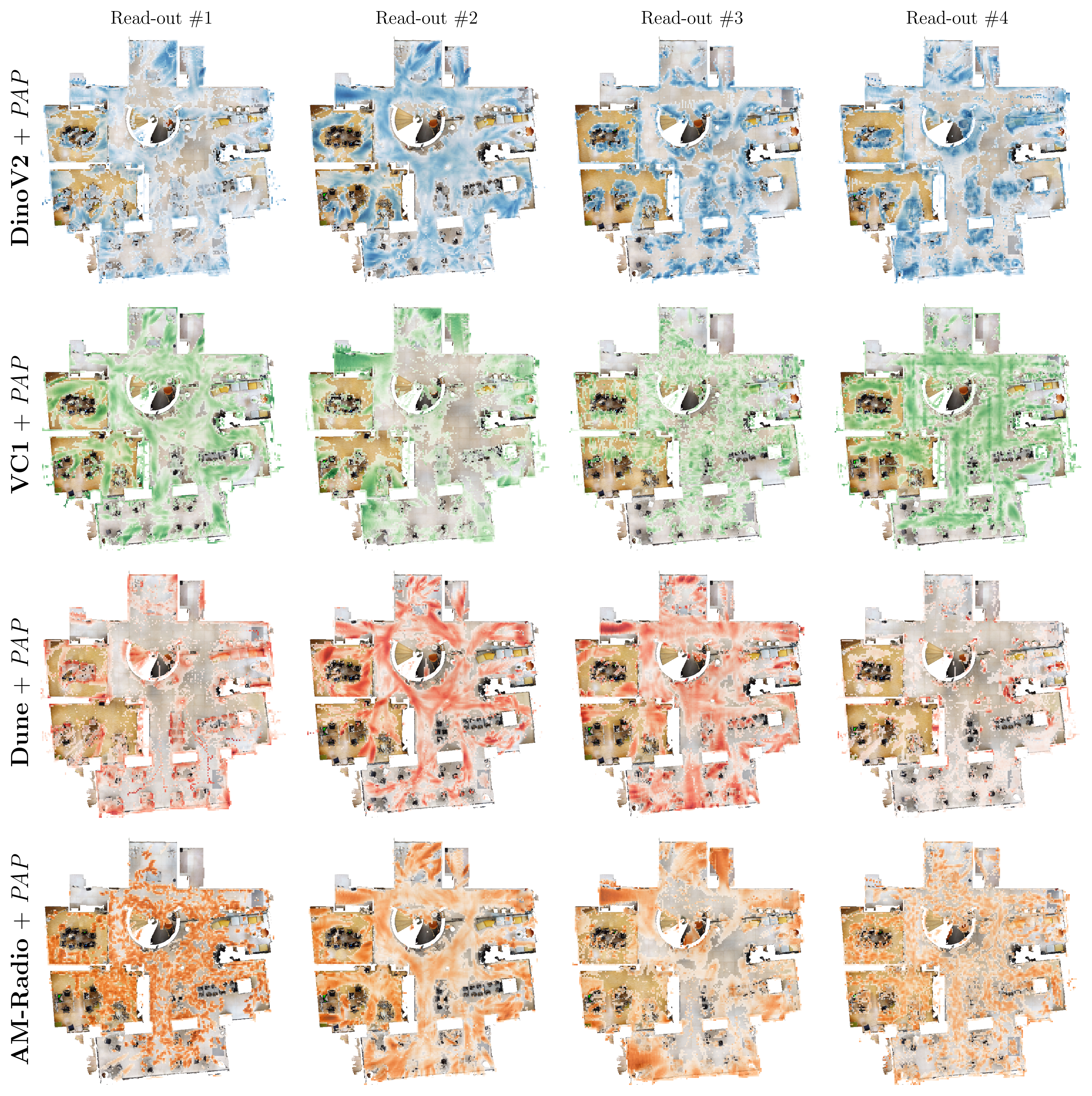}
    \caption{\label{fig:accummaps}\textbf{Attention maps} -- with pure attention projection (PAP) accumulated and superimposed over a map. %
    \simbox{(data collected in simulation)}.}
    \vspace*{-4mm}
\end{figure}

\myparagraph{Interpretable spatial maps emerge from pre-trained encoders.} All previously discussed experiments have been performed with Cross-attention pooling (CAP). We now introduce a new pooling, ``\textit{Pure Attention Projection (PAP)}'' described at the beginning of this section. It is derived from Cross-attention pooling, completely discards the value projections and has the peculiar (and as we argue, very useful) property that the entirety of the features passed to the downstream agent corresponds to interpretable heatmaps. In other words, compared to other methods for interpretability in machine learning, there is no additional hidden information passed to the model, which a human would not see when inspecting interpretable maps. This is achieved while maintaining downstream navigation performance (see Table \ref{tab:ablationnumro}, commented in detail further below).

Similar to CAP, PAP uses a small set of trained query tokens $\mathbf{q}_n$ and produces an attention map per token. Figure \ref{fig:affordances} shows these maps overlayed over a real input image for 4 different pre-trained ViTs and for a single selected read-out token out of 4 tokens. We have observed that particular semantic meanings emerge in these maps, and found particularly exciting the emergence of what we call ``\textit{affordance map}'' in this context: attention seems to focus on the areas in the scene, where an agent could potentially go when the current position / area needs to be left. Note, that this is an ``unconditional'' perception module, it is not conditioned on the agent's goal. We also found it very interesting that this behavior emerges in projection layers for every pre-trained ViT we evaluated. We would like to stress again, that the heatmaps shown in Fig. \ref{fig:affordances} (one map for each of the 4 read-out tokens) are \textbf{the only information available to the agent:} there are no additional embeddings or features, except odometry input, which does not provide information on the scene.

Fig. \ref{fig:sim2real} shows that these maps are robust wrt. to the sim2real gap. The attention on a simulated region in the scene is similar to the attention on a similar input from the real robot's camera. For convenience we overlayed arrows indicating the same regions in our test building. 

In Fig. \ref{fig:accummaps} we investigate how these heatmaps are linked to scene structure and semantics. We un-project and spatially integrate them into a global map, which we overlay over a map in simulation. We observed that they correspond to emerged concept detectors linked to navigable space, obstacles and affordances.

\begin{figure}[t!]
    \centering
    \includegraphics[width=0.49\linewidth]{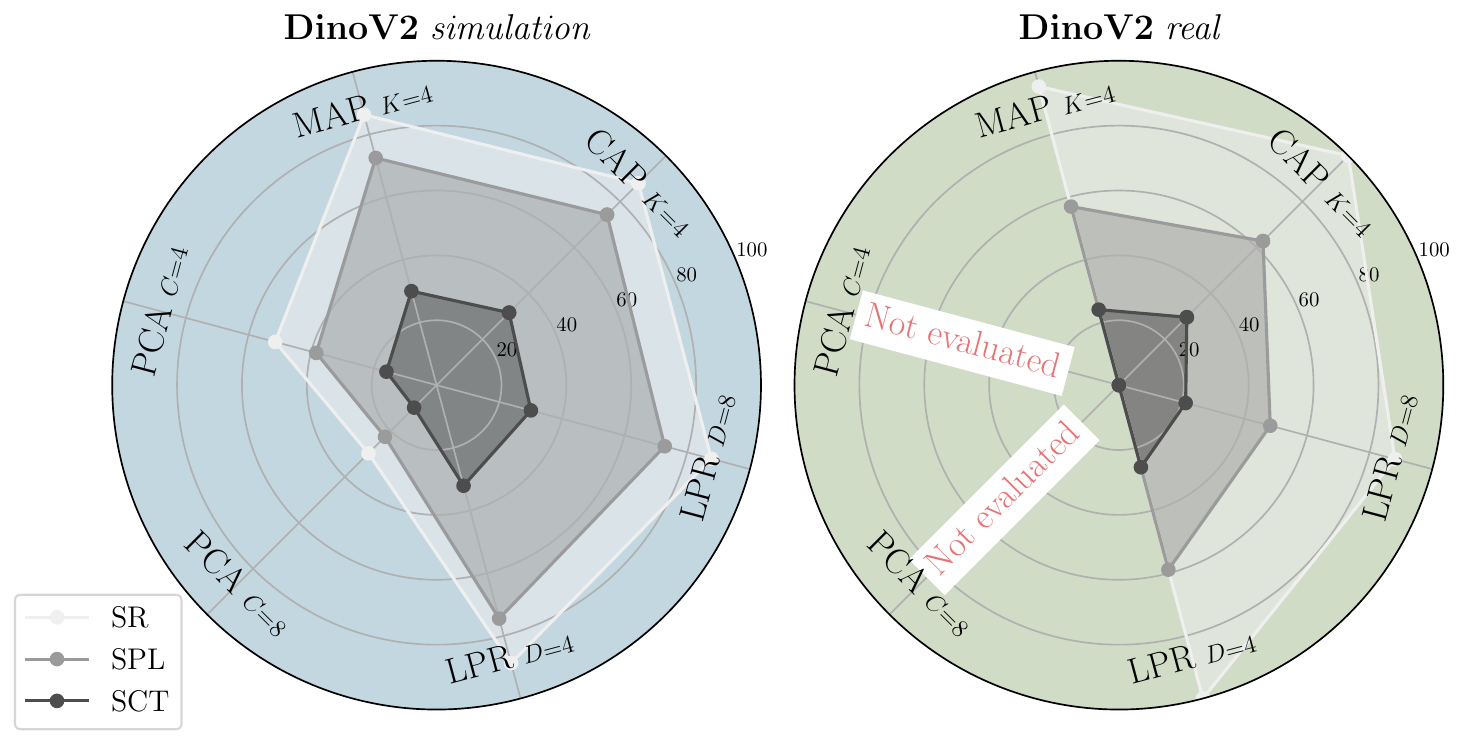}
    \hfill
    \includegraphics[width=0.49\linewidth]{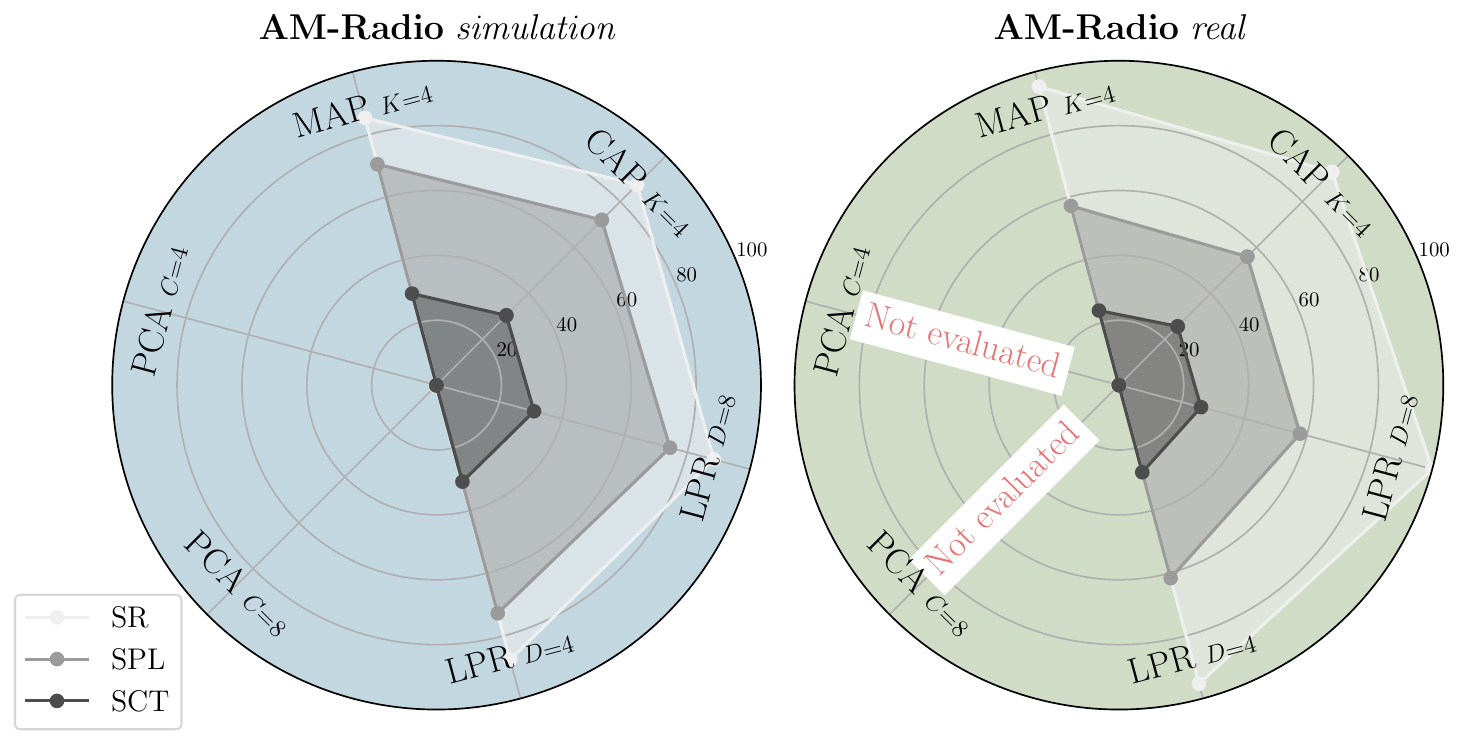}
    \caption{\label{fig:projectors}\textbf{Comparing different projection mechanisms} -- applied to DinoV2 and AM-Radio encoders. Models are initialized from privileged agent and fine-tuned for 100M steps. We observe similar or better performances in sim and real with projectors that bottleneck with low complexity (cf. Sec. \ref{sec:spatialityprojectors}).    
    See Table \ref{tab:projectors} in the \suppmat{} for numbers and variations, ie. avg${\pm}$std-dev \realbox{(336 real nav episodes in this Figure)}.}    
\end{figure}

\begin{table}[t]
    \setlength{\aboverulesep}{0pt}
    \setlength{\belowrulesep}{0pt}
    \centering
    \begin{tabularx}{\textwidth}{ll| HHH || RRR }
        \myrule
        \rowcolor{darkgray}ViT & & 
        \multicolumn{3}{c||}{\cellcolor{coolblue!40}Sim} & 
        \multicolumn{3}{c}{\cellcolor{coolgreen!40}Real} \\ 
        \cellcolor{darkgray}Backbone & \multirow{-2}{*}{\cellcolor{darkgray}Projector} &  
        \cellcolor{coolblue!30}\scriptsize{SCT}  & \cellcolor{coolblue!30}\scriptsize{SPL}  & \cellcolor{coolblue!30}\scriptsize{SR}  & 
        \cellcolor{coolgreen!30}\scriptsize{SCT} & \cellcolor{coolgreen!30}\scriptsize{SPL} & \cellcolor{coolgreen!30}\scriptsize{SR} \\ \myrule
        \cellcolor{midgray}                           &\cellcolor{lightgray} $K{=}1$  & 31.3 & 74.8 & 88.0 & 28.8 \textmuted{$\pm 0.6$} & 61.7 \textmuted{$\pm 1.5$} & \textbf{97.6} \textmuted{$\pm 3.3$} \\ 
        \cellcolor{midgray}                           &\cellcolor{lightgray} $K{=}2$  & \textbf{32.0} & \textbf{75.2} & \textbf{89.7} & 23.9 \textmuted{$\pm 1.4$} & 54.6 \textmuted{$\pm 2.1$} & 95.2 \textmuted{$\pm 6.7$} \\ 
        \cellcolor{midgray}                           &\cellcolor{lightgray} $K{=}4$  & 31.6 & 74.3 & 88.0 &  \textbf{27.5} \textmuted{$\pm 4.2$} & \textbf{57.2} \textmuted{$\pm 9.4$} & \textbf{97.6} \textmuted{$\pm 3.3$} \\ 
        \multirow{-4}{*}{\cellcolor{midgray}\shortstack{DinoV2 +\\ CAP}} 
                                                      & \cellcolor{lightgray} $K{=}8$ & 30.8 & 73.2 & 88.6 & 22.6 \textmuted{$\pm 2.0$} & 53.1 \textmuted{$\pm 4.5$} & 90.5 \textmuted{$\pm 6.7$}  \\ \midrule
        \cellcolor{midgray}                           &\cellcolor{lightgray} $K{=}1$  & 25.3 & 66.5 & 81.7 & 20.4 \textmuted{$\pm 0.3$} & 54.9 \textmuted{$\pm 1.4$} & 92.9 \textmuted{$\pm 0.0$}  \\ 
        \cellcolor{midgray}                           &\cellcolor{lightgray} $K{=}2$  & 29.2 & 72.2 & 86.8 & 23.2 \textmuted{$\pm 2.4$} & 55.8 \textmuted{$\pm 2.2$} & \textbf{100} \textmuted{$\pm 0.0$}  \\ 
        \cellcolor{midgray}                           &\cellcolor{lightgray} $K{=}4$  & 30.0 & 72.5 & 86.3 &  \textbf{24.1} \textmuted{$\pm 1.9$} & \textbf{57.0} \textmuted{$\pm 3.5$} & 95.3 \textmuted{$\pm 3.3$} \\ 
        \multirow{-4}{*}{\cellcolor{midgray}\shortstack{DinoV2 +\\ PAP}} 
                                                      & \cellcolor{lightgray} $K{=}8$ & \textbf{30.5} & \textbf{75.1} & \textbf{89.3} & 23.5 \textmuted{$\pm 1.3$} & 53.9 \textmuted{$\pm 0.7$} & 92.9 \textmuted{$\pm 0.0$} \\ \myrule
    \end{tabularx}
    \caption{\label{tab:ablationnumro}\textbf{Sensitivity analysis of number of read-out tokens} -- for CAP and PAP projectors. While performance in simulation tends to increase with the number of read-out tokens, this does not transfer well on real experiments. Agent finetuned from Lidar (comparable to Table \ref{tab:from_mibi}). \realbox{(336 real nav episodes in this table)}.}
    \vspace{-8mm}
\end{table}

\myparagraph{How much visual information is needed?} We have seen that passing simple attention maps to the downstream agent leads to interpretable features, and on the spider plots in Figure \ref{fig:projectors} we see that this does not sacrifice downstream navigation performance. We evaluate the different pooling methods introduced in the beginning of this section, CAP, PAP, PCA and LPR, using DinoV2 and AM-Radio vision encoders for evaluation in \simbox{simulation} and \realbox{real} (336 real navigation episodes have been performed in our test building for this figure alone). PCA does surprisingly underperform, we observe similar or better performances in simulation and real with projectors that conserve spatial information (full data including variations w. avg $\pm$ std-dev is given in Table \ref{tab:projectors} in the \suppmat{}).

This raises the question how much information from each patch embedding delivered by the ViTs is actually useful and \textit{required} for downstream agents. Table \ref{tab:ablationnumro} provides a sensitivity analysis, where we vary the number $K$ of read-out tokens in the set $\{1,2,4,8\}$ for both CAP and PAP projectors. For CAP, given $K$ read-out tokens, this leads to $K{\cdot}E$ values, where $E$ is the embedding dimension of the ViT. For PAP, this leads to $K{\cdot}P$ values, where $P$ is the amount of patches in the input image, each heatmap being composed of $P$ values. \textbf{We observe that excellent navigation performance even in real experiments can be achieved with as low as 1 scalar per image patch passed to the downstream agent.} While performance in simulation tends to increase with the number of read-out tokens, this does not transfer well to real experiments, corroborating that bottlenecking has a positive impact on the sim2real gap.

\begin{table}[t]
    \centering
    \setlength{\aboverulesep}{0pt}
    \setlength{\belowrulesep}{0pt}
    \begin{tabularx}{\textwidth}{l| HHH || RRR}
        \myrule
        \rowcolor{darkgray} &
        \multicolumn{3}{c||}{\cellcolor{coolblue!40}Sim} & 
        \multicolumn{3}{c|}{\cellcolor{coolgreen!40}Real} \\ 
        \cellcolor{darkgray} \multirow{-2}{*}{Setup} &  
        \cellcolor{coolblue!30}\scriptsize{SCT}  & \cellcolor{coolblue!30}\scriptsize{SPL}  & \cellcolor{coolblue!30}\scriptsize{SR}  & 
        \cellcolor{coolgreen!30}\scriptsize{SCT} & \cellcolor{coolgreen!30}\scriptsize{SPL} & \cellcolor{coolgreen!30}\scriptsize{SR} \\ \myrule
        \cellcolor{midgray} CAP \scriptsize{$K{=}4$}, Seed \#1   & 00.0 & 00.0 & 00.0 & & & \\ 
        \cellcolor{midgray} CAP \scriptsize{$K{=}4$}, Seed \#2  & 00.0 & 00.0 & 00.0 & & &\\ 
        \cellcolor{midgray} CAP \scriptsize{$K{=}4$}, Seed \#3   & 00.0 & 00.0 & 00.0 & \multicolumn{3}{c}{\multirow{-3}{*}{\cellcolor{coolgreen!10}{\textit{not evaluated}}} }\\  \midrule

        \cellcolor{midgray} CAP \scriptsize{$K{=}2$}   & 00.0 & 00.0 & 00.0 & & & \\ 
        \cellcolor{midgray} CAP \scriptsize{$K{=}8$}  & 00.0 & 00.0 & 00.0 & \multicolumn{3}{c}{\multirow{-2}{*}{\cellcolor{coolgreen!10}{\textit{not evaluated}}}}\\ \midrule

        \cellcolor{midgray} CAP \scriptsize{$K{=}4$}, ViT/s & 00.0 & 00.0 & 00.0 &\multicolumn{3}{c}{\cellcolor{coolgreen!10}{\textit{not evaluated}}}\\  \midrule
        
        \cellcolor{midgray} PAP \scriptsize{$K{=}2$} & 12.8 & 41.6 & 52.1 & \multicolumn{3}{c}{\cellcolor{coolgreen!10}{\textit{not evaluated}}}\\
        \cellcolor{midgray} PAP \scriptsize{$K{=}4$} & 16.1 & 47.0 & 59.8 & \multicolumn{3}{c}{\cellcolor{coolgreen!10}{\textit{not evaluated}}} \\
        \cellcolor{midgray} PAP \scriptsize{$K{=}8$} & \textbf{24.9} & \textbf{63.7} & \textbf{79.7} & \textbf{24.3} \textmuted{$\pm 0.2$} & \textbf{58.1} \textmuted{$\pm 1.8$} &\textbf{95.3} \textmuted{$\pm 3.3$} \\
        \cellcolor{midgray} LPR \scriptsize{$D{=}4$} & 21.7 & 59.6 & 77.7 & 21.0 \textmuted{$\pm 2.1$} & 49.6 \textmuted{$\pm 2.6$} & 88.1 \textmuted{$\pm 3.4$} \\
        \cellcolor{midgray} LPR \scriptsize{$D{=}8$} & 10.2 & 23.6 & 31.6 & \multicolumn{3}{c}{\cellcolor{coolgreen!10}{\textit{not evaluated}}} \\ \myrule
    \end{tabularx}
    \caption{\label{tab:dinov3}\textbf{DinoV3 required extra efforts} compared to the other ViTs. We tested different configurations and seeds for CAP, but failed systematically. However, less complex, spatial preserving projectors like PAP and LPR manage to exploit DinoV3 features while transferring well to the real world. \realbox{(84 real nav episodes in this table)}.} 
    \vspace*{-8mm}
\end{table}

\myparagraph{The special case of DinoV3}. The extremely bad performance of DinoV3 (cf. Tables \ref{tab:from_scratch} and \ref{tab:from_mibi}) was surprising given the gains it achieved over its predecessor DinoV2 over a set of standard CV tasks. We particularly investigated this to exclude any potential mistakes in the setup. Table \ref{tab:dinov3} shows that the classical CAP projector completely failed for this ViT even when exploring different numbers of read-out tokens, changing the size from ViT/B to ViT/S, and of course training with different seeds. However, it was interesting that we could make DinoV3 work with the new interpretable PAP projector. We conjecture that the complexity is at stake, and CAP fails to identify required features from DinoV3 from the weak reward signal. On the other hand, PAP and LRP --while less expressive -- are also less complex and preserve spatiality of the representation.

These findings are somewhat corroborated by other experimental studies. As found in \cite{huang2025rethinkingforgery}, DinoV3 relies on globally coherent low-frequency structures, tends to ignore high-frequency components, and ``\textit{DinoV3’s decision boundary critically depends on maintaining global spatial organization rather than on localized details}''. Similarly, a study for medical image analysis \cite{liu2026dinov3medical} showed that DinoV3’s natural image features do not transfer to the fine-grained textural analysis. 
\vspace*{-2mm}

\section{Conclusion}
We presented a large-scale study of visual pre-training strategies for real-world navigation and  conducted \numepisodes{} episodes with a real robot in a real building.
We show that while learning perception directly from actions alone fails to generalize, leveraging state-of-the-art encoders, and particularly those distilled from heterogeneous teachers like AM-Radio \cite{radio25} and Dune \cite{dune2025}, enables robust sim-to-real transfer when combined with realistic motion modelling. We introduced a new projection layer called ``\textit{Pure Attention Projection}'' which reveals that navigational information can be extracted from frozen ViTs and compressed into highly efficient, interpretable bottleneck representations that naturally encode affordances without loss of performance. Our results suggest that real-world navigation can be achieved by training agents entirely in simulation if  perception is delegated to rich vision encoders combined with proper projections and that robot motion is simulated realistically during training.

\bibliographystyle{splncs04}
\bibliography{one_bib_please.bib}
\newpage
\appendix
\begin{center}
    \large
    \textbf{Supplementary Material for Paper \#1424 \\ ~\\   
    \savedtitle
    }
\end{center}

\section{Detailed agent architecture and training}

\subsection{Agent architecture}
Table \ref{tab:lineupvits} reports additional information about the ViTs used throughout the paper. The on-board camera captures $160\times 120$ images at \SI{3}{\hertz}, matching the sampling frequency of the agent. The images are zero-padded and up-scaled to match the resolution expected by the visual encoder. We extract dense features via the frozen ViT in half-precision mode during both training and inference. This guarantees real-time execution on the Jetson Orin running the model on the real platform. Unless explicitly stated in the main paper, we use the ViT-Base version of the pretrained encoders. 

\vspace{2mm}
The projector is directly applied to the dense features (patch + registers embeddings) given by the visual encoder. Let $\bF_p \in \RR^E$ be this output, where $p \in \llbracket 0, P\rrbracket$ is the number of patch embeddings.
\begin{description}
    \item[Cross Attention Pooling] (CAP) perform cross multi-head attention (MHA with 4 heads) between the $P$ embeddings from the ViT backbone and $K$ learnable features $\bmq_k \in \RR^{E}$. The MHA outputs $K$ latent embedding of size $E$, which are concatenated and projected to a final dimension of $1024$ with a linear layer.

    \item[Pure Attention Projection] (PAP) directly returns the attention weights from the cross attention, and ignore value projection. Concretely, each pair of patch embedding $\bF_p$ and learnable read-out token $\bmq_k$ gives a single scalar value corresponding to the cosine similarity between $\bF_p$ and $\bmq_k$, scaled by \textit{softmax}. We obtain $K$ latent vectors of size $P$, which are concatenated and project to a final dimension of $1024$ with a linear layer. 

    \item[Principal Component Analysis] (PCA) is frequently used to visualize dense embeddings from pre-trained ViT (by computing the top-3 principal component, converted to RGB channels). We built on this idea to see if static projection is sufficient for navigation (a negative result, see table \ref{tab:projectors}). We initialize the top $C$ principal components (PCs) from 1M images collected during navigation in simulation. During training, each patch is projected independently onto the PCs. The projection weights are stacked patch-wise and projected to a final dimension of $1024$ with a linear layer.

    \item[Linear Patchwise Reduction] (LPR) projects each patch embedding independently with a linear layer, reducing their dimension from $E$ to $D$ (typically $D{=}4$ or $D{=}8$ scalar values). The resulting vectors are stacked patch-wise and projected to a final dimension of $1024$ with a linear layer. 
\end{description}

\begin{table}[t] \centering
    \setlength{\aboverulesep}{0pt}
    \setlength{\belowrulesep}{0pt}
    \begin{tabularx}{\textwidth}{l|c|X|c}
        \myrule      
         \rowcolor{midgray} ViT & \# Params. & Pre-training & img. size  \\ \myrule
         \multicolumn{4}{c}{\textit{From scratch}} 
         \\
         \midrule
         \cellcolor{lightgray}Dino-v2 \cite{dinov22024}    & 86.23M & Self-distillation \cite{dino2021} + iBOT \cite{ibot2022} & $224{\times} 224$ \\
         \cellcolor{lightgray}Dino-v3 \cite{dinov32025}   & 85.67M& Self-distillation \cite{dino2021} + iBOT \cite{ibot2022} & $224{\times}224$ \\ 
         \cellcolor{lightgray}VC-1 \cite{vc12024} & 85.80M& MAE \cite{mae2022} on egocentric images & $224{\times}224$\\ \midrule
         \multicolumn{4}{c}{\textit{Distilled}} 
         \\ 
         \midrule
         \cellcolor{lightgray}DUNE \cite{dune2025}      & 85.97M & Dino-V2 \cite{dinov22024}, Mast3R \cite{mast3r}, Multi-HMR \cite{mhmr2024} & $336{\times}336$ \\ 
         \cellcolor{lightgray}AM-Radio v2.5 \cite{radio25} & 98.23M & CLIP (DFN\cite{fang2024data} and OpenAI \cite{radford2021learning}), Dino-V2\cite{dinov22024}, SAM\cite{kirillov2023segment} & $144{\times}256$ \\ \myrule
    \end{tabularx}
    \caption{\label{tab:lineupvits}\textbf{Pre-trained ViT models} -- compared throughout this paper. We used the \textit{base} version of each visual encoders as it offers a good trade-off between performances, training speed and real-time inference time.}
\end{table}

The agent maintains a latent memory modelled by a 2-layer gated recurrent unit. The latent memory size is $1024$ and is updated from latent observations:
\begin{itemize}
    \item \textbf{RGB} is given by the output of the chosen projector, of size $1024$,
    \item \textbf{Odometry} sensor returns the 2D position and orientation of the robot with respect to the episode start, as well as linear and angular velocities. The raw sensor values are encoded by a  1-layer MLP up to a final size of $64$,
    \item \textbf{Previous action} is mapped to a learnable embedding of size $32$,
    \item \textbf{Static point goal} is encoded with a 1-layer MLP up to a final size of $64$.
\end{itemize}

The policy is a linear layer applied on the latent memory of the agent. The pretrained agent on privileged observations (used in table \ref{tab:from_mibi} does not have an RGB encoder, but instead encodes the 2D Lidar observation with a 1D CNN.

\subsection{Training and evaluation setup}
\begin{table}[t]
    \centering
    
    \begin{tabularx}{\textwidth}{ll| HHH || RRR }
        \myrule
        \rowcolor{darkgray}ViT & & 
        \multicolumn{3}{c||}{\cellcolor{coolblue!40}Sim} & 
        \multicolumn{3}{c}{\cellcolor{coolgreen!40}Real} \\ 
        \cellcolor{darkgray}Back. & \multirow{-2}{*}{\cellcolor{darkgray}Projector} &  
        \cellcolor{coolblue!30}\scriptsize{SCT}  & \cellcolor{coolblue!30}\scriptsize{SPL}  & \cellcolor{coolblue!30}\scriptsize{SR}  & 
        \cellcolor{coolgreen!30}\scriptsize{SCT} & \cellcolor{coolgreen!30}\scriptsize{SPL} & \cellcolor{coolgreen!30}\scriptsize{SR} \\ \myrule
        \cellcolor{midgray}                           &\cellcolor{lightgray} CAP   & 31.6& 74.3 & 88.0 & 27.5 \textmuted{$\pm 4.2$} & 57.2 \textmuted{$\pm 9.4$} & 97.6 \textmuted{$\pm 3.3$} \\ 
        \cellcolor{midgray}                           &\cellcolor{lightgray} PAP      & 30.0 & 72.5 & 86.3 & 24.1 \textmuted{$\pm 1.9$} & 57.0 \textmuted{$\pm 3.5$} & 95.3 \textmuted{$\pm 3.3$} \\ 
        \cellcolor{midgray}                           &\cellcolor{lightgray} LPR \scriptsize{$D{=}4$}      & \textbf{32.1} & \textbf{74.5} & \textbf{88.7} & 26.2 \textmuted{$\pm 0.4$} & 58.9 \textmuted{$\pm 0.4$} & \textbf{100} \textmuted{$\pm 0.0$} \\
        \multirow{-6}{*}{\cellcolor{midgray}DinoV2}   &\cellcolor{lightgray} LPR \scriptsize{$D{=}8$}     & 30.1 & 71.8 & 87.7 & 21.3 \textmuted{$\pm 2.9$} & 48.3 \textmuted{$\pm 4.4$} & 88.1 \textmuted{$\pm 6.7$} \\ 
        \cellcolor{midgray}                           &\cellcolor{lightgray} PCA  \scriptsize{$C{=}4$}      & 16.0 & 38.4 & 51.5 & \multicolumn{3}{c}{\cellcolor{coolgreen!10}{\textit{not evaluated}}}  \\ 
        \cellcolor{midgray}                           &\cellcolor{lightgray} PCA \scriptsize{$C{=}8$}      & 9.8 & 22.5 & 29.7 & \multicolumn{3}{c}{\cellcolor{coolgreen!10}{\textit{not evaluated}}}  \\
        \midrule
        \cellcolor{midgray}                             &\cellcolor{lightgray} CAP    & 30.4 & 72.0 & 87.4 & 25.6 \textmuted{$\pm 0.7$} & 56.0  \textmuted{$\pm 1.5$} & 92.9  \textmuted{$\pm 5.8$}  \\ 
        \cellcolor{midgray}                             &\cellcolor{lightgray} PAP      & 29.2 & 70.5 & 85.3 & 23.8 \textmuted{$\pm 1.3$} & 57.2  \textmuted{$\pm 3.3$} & 95.3  \textmuted{$\pm 3.3$} \\ 
        \cellcolor{midgray}                             &\cellcolor{lightgray} LPR \scriptsize{$D{=}4$}     & 30.8 & 72.8 & 87.6 & \textbf{27.1} \textmuted{$\pm 2.0$} & \textbf{60.6} \textmuted{$\pm 3.9$} & 95.3  \textmuted{$\pm 3.3$} \\
        \multirow{-6}{*}{\cellcolor{midgray}AM-Radio}   &\cellcolor{lightgray} LPR \scriptsize{$D{=}8$}     & \textbf{31.1} & \textbf{74.5} & \textbf{88.5} & 26.2  \textmuted{$\pm 2.8$} & 58.0  \textmuted{$\pm 6.4$} & \textbf{100}  \textmuted{$\pm 0.0$}\\ 
        \cellcolor{midgray}                             &\cellcolor{lightgray} PCA \scriptsize{$C{=}4$}      & 0 & 0 & 0 & \multicolumn{3}{c}{\cellcolor{coolgreen!10}{\textit{not evaluated}}} \\
        \cellcolor{midgray}                             &\cellcolor{lightgray} PCA \scriptsize{$C{=}8$}      & 0.1 & 0.2 & 0.4 & \multicolumn{3}{c}{\cellcolor{coolgreen!10}{\textit{not evaluated}}}  \\
        \myrule
    \end{tabularx}
    \caption{\textbf{Ablation on ViT projection} --  Detailed metrics associated with figure \ref{fig:projectors}. All variants are initialized from the pre-trained agent with privileged information and finetuned for 100M steps. In general, we found beneficial to preserve spatiality in the final representation.}
    \label{tab:projectors}
\end{table}

As mentioned in the main paper, each agent is trained with reinforcement learning, in particular PPO~\cite{schulman2017proximal}, and with a reward inspired by \cite{chattopadhyay2021robustnav,janny2025},
$
r_t=\mathrm{R} \cdot \mathbb{I}_{\text {success}} -\Delta_t^{\mathrm{Geo}} -\lambda -\mathrm{C} \cdot \mathbb{I}_\text{collision}
$, where $R{=}2.5$, $\Delta_t^{\mathrm{Geo}}$ is the gain in geodesic distance to the goal, a slack cost $\lambda{=}0.01$ encourages efficiency, and a cost $C=0.1$ penalizes each collision without ending the episode. Models from scratch (in table \ref{tab:from_scratch}) are trained for 500M steps on a single A100 GPU ($\sim$ 3-4 weeks of training), and agent derived from privileged agent are finetuned for 100M steps. The privileged agent has been trained for 800M steps, following \cite{janny2025reasoning}.

We evaluate each models in both simulated and real environment. For security reasons and to prevent dangerous situation, we excluded from real world experiment any model that performs below 60\% in success rate in simulation. We repeated each sequence of real episodes $\times 3$ to account for uncertainty and changing environmental condition, such as lighting. An episode is considered successful if the agent stops within \SI{0.2}{\meter} of the goal in simulation (episode is terminated at first stop), and within \SI{1}{\meter} in real. Additionally, we found beneficial to perform an agent reset when approaching \SI{2}{\meter} of the goal, based on local odometry measurements. In that case, the internal state is reset and the point goal is fed again to the agent. Moreover, the robot integrates a safety mechanism that prevent collision with object and people based on obstacle detection from an on-board 2D Lidar. While this sensory input is \textit{not} provided to the agent, the safety mechanism is simulated during training.

\section{Additional experiments}

Table \ref{tab:projectors} reports metrics displayed in figure \ref{fig:projectors} in the main paper. Our proposed alternatives for patch embedding projection performs on par or better than classical cross-attention projection. PAP and LPR methods are arguably less expressive than CAP, yet still enable efficient navigation in both simulated and real environments. 

\begin{figure}[t]
    \centering
    \includegraphics[width=\linewidth]{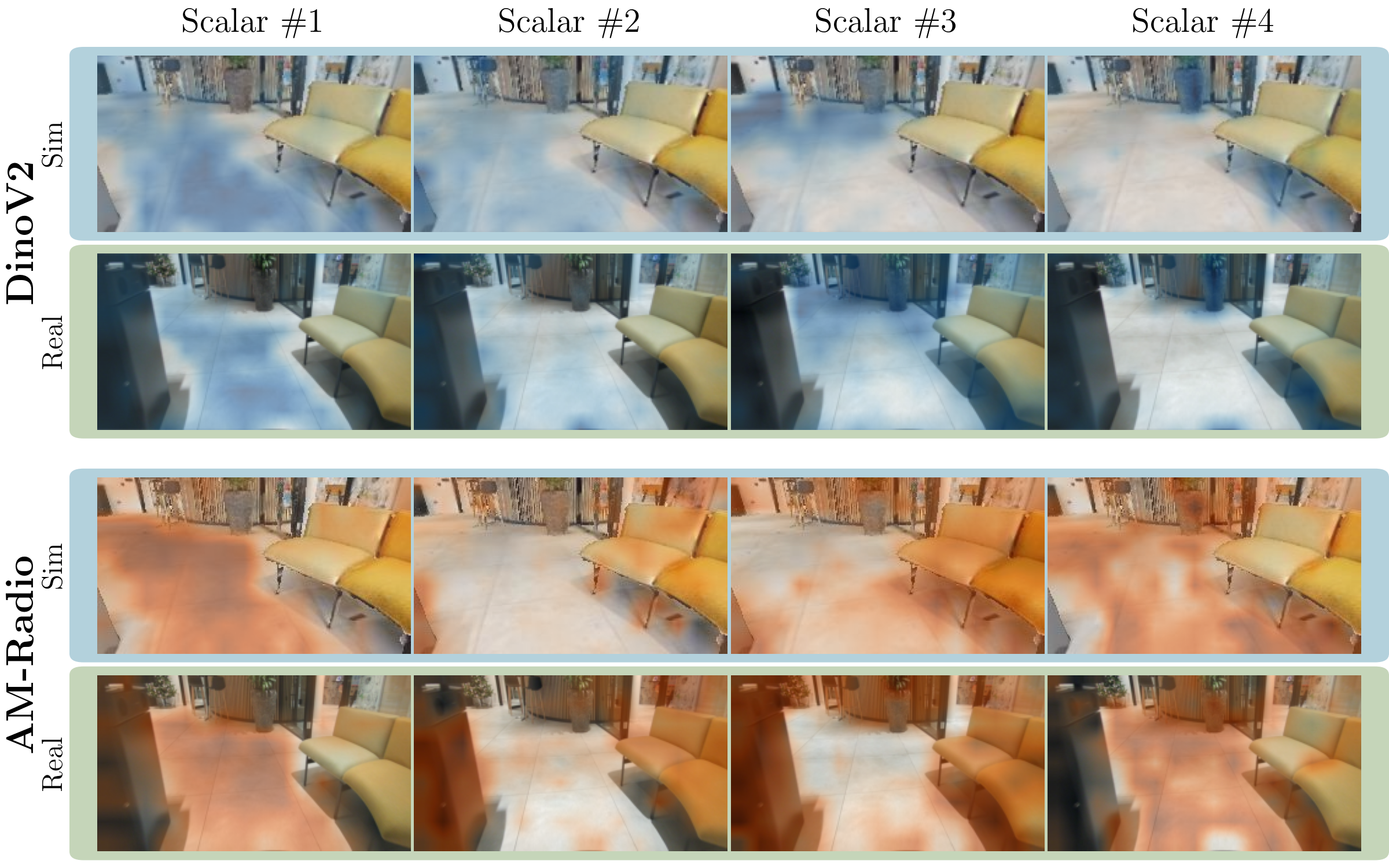}
    \caption{\textbf{Magnitude from LPR} -- can be seen akin to attention maps from PAP, and exhibits very similar behavior. Sim2real gap is limited, as corresponding region in simulated and real images trigger similar magnitude level.}
    \label{fig:lpr_explain}
\end{figure}

Remarkably, we observe very similar explainable feature maps with PAP and LPR, which indicates that the detected semantic classes are fundamental for navigation. We recall that these maps emerge solely from RL training and are not supervised directly. Figure \ref{fig:lpr_explain} shows the magnitude of the intermediate representations from LPR, and exhibit segmentation-like properties for navigable floor, obstacles and affordances.

\subsection{Cross-attention projection(CAP)}
\begin{figure}[t]
    \centering
    \includegraphics[width=\linewidth]{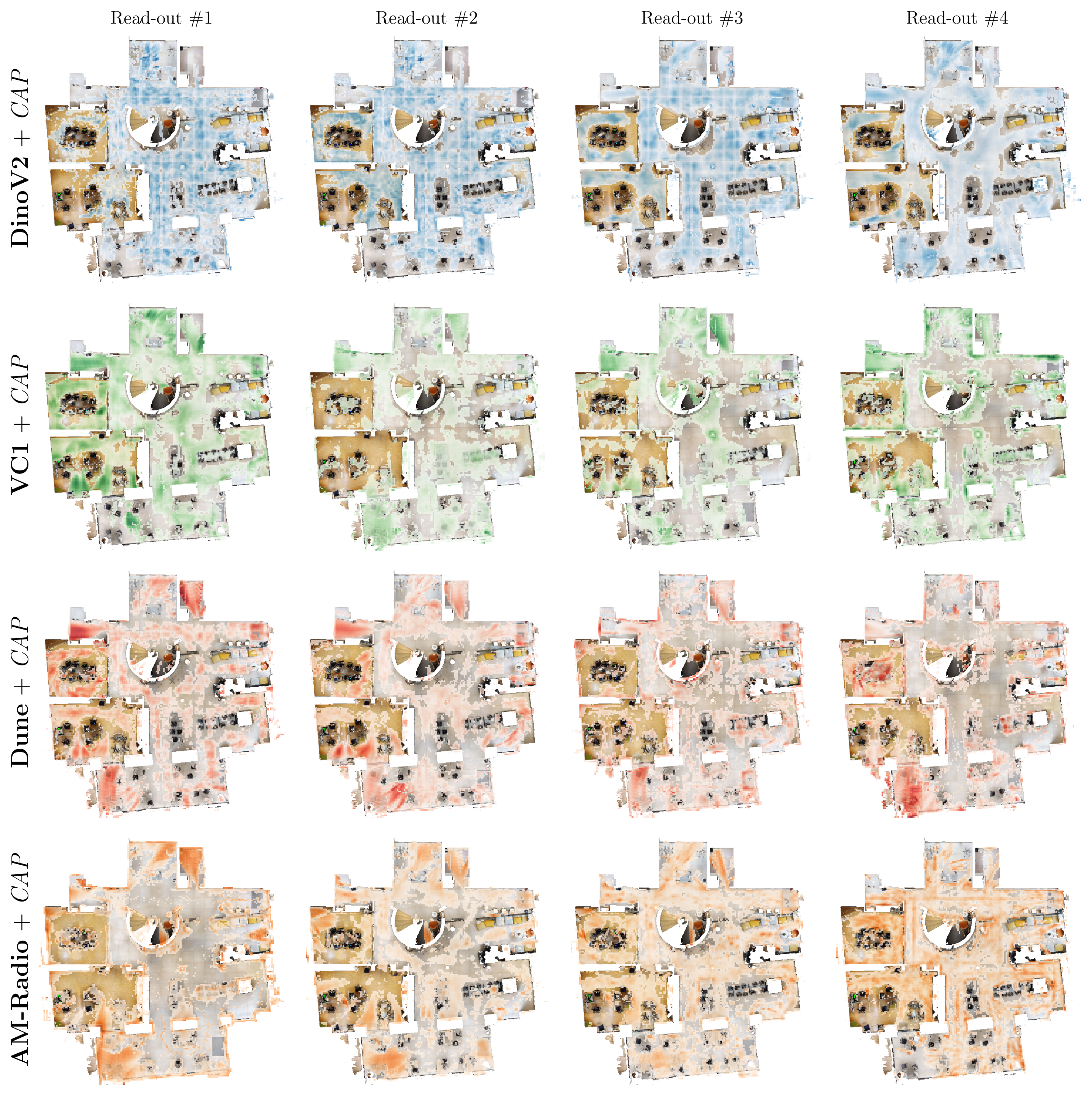}

    \caption{\label{fig:accummaps_cap}\textbf{Attention maps} -- with cross attention projection (CAP) accumulated and superimposed over a map. %
    \simbox{(data collected in simulation)}.}
    \vspace*{-4mm}
\end{figure}

We also provided projected attention maps derived from CAP (using models from table \ref{tab:from_mibi}, i.e. initialized from privileged agent). Conversely to PAP, the attention maps are not clearly interpretable. We argue that most of the information is stored in the value projection, which can't be visualized easily. Surprisingly, DinoV2 with CAP exhibits a grid-like pattern which aligns with the floor tiling of our test environment. 

\begin{figure}[t]
    \centering
    \begin{tikzpicture}[font=\footnotesize]
        \draw (0, 0) node[anchor=west, inner sep=0] {\includegraphics[width=\textwidth]{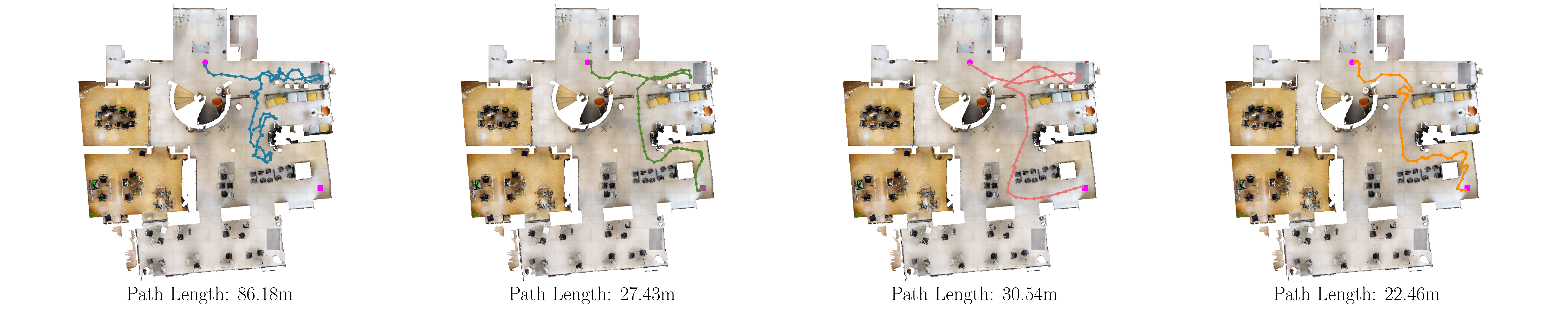}};
        \draw (0, -2.4) node[anchor=west, inner sep=0] {\includegraphics[width=\textwidth]{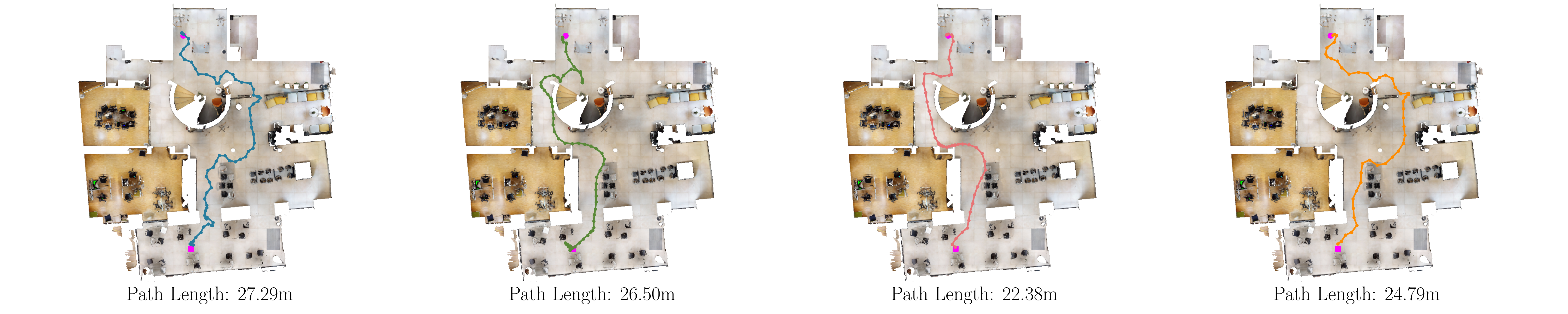}};
        \draw (0, -4.8) node[anchor=west, inner sep=0] {\includegraphics[width=\textwidth]{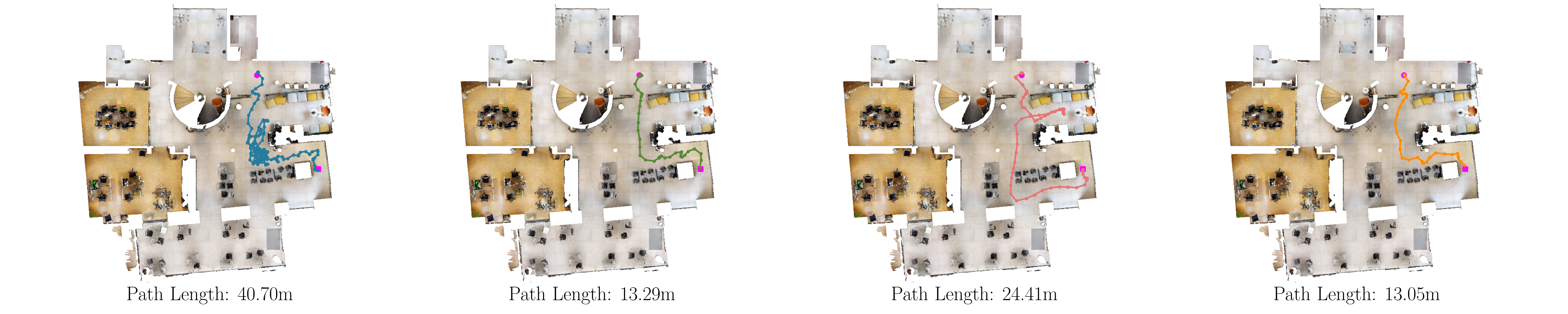}};
        \draw (0.13\textwidth, 1.4) node[anchor=center, inner sep=0] {\textbf{DinoV2}};
        \draw (0.38\textwidth, 1.4) node[anchor=center, inner sep=0] {\textbf{VC1}};
        \draw (0.62\textwidth, 1.4) node[anchor=center, inner sep=0] {\textbf{Dune}};
        \draw (0.86\textwidth, 1.4) node[anchor=center, inner sep=0] {\textbf{AM-Radio}};
    \end{tikzpicture}
    \caption{\textbf{Sampled trajectories from \textreal{real experiments}} -- using models trained from scratch (table \ref{tab:from_scratch}). Overall, distilled models show better trajectories closer from the optimal path, while exhibiting better behavior with less hesitation and backtracking. For each model and episode, we pick the realization with the highest SPL. The agent is tasked to navigate from \textcolor{magenta}{\LARGE{$\bullet$}} to \textcolor{magenta}{$\blacksquare$}.}
    \label{fig:real_traj_scratch}
\end{figure}

Figure \ref{fig:real_traj_scratch} show trajectory samples collected during real episodes with agents trained from scratch. Each figure shows the best realization for a given episode per model among the 3 repetitions. While the robot reaches the goal most of the time, the trajectory can be quite far from the optimal shortest path, which reflects performances measured in simulation. Qualitatively, \textit{DinoV2} trajectories contain more spurious rotations which makes displacement erratic.

\subsection{Investigating the impact of spatiality for projectors}
\label{sec:spatialityprojectors}

The failure of \textit{DinoV3} under several variants of projectors was a surprise, but we manage to obtain encouraging results with LPR and PAP, which led us to two potential explanations. We believe that extracting simple information for navigation from \textit{DinoV3}'s complex features might be too difficult, especially from pure RL training. We hypothesize that the relative success of LPR and PAP might be explained by either (1) their ability to preserve spatial information in the representation, or (2) by the simplicity and weaker expressivity of the projector itself, which could pair better with reinforcement learning.

\begin{wrapfigure}{r}{0.5\textwidth}
    \centering
    \setlength{\tabcolsep}{0.5em}
    \begin{tabularx}{0.5\textwidth}{l| HHH  }
        \myrule
        \multicolumn{4}{c}{\textit{Self-attention pooling}} \\
        \myrule
        \cellcolor{darkgray}ViT & \multicolumn{3}{c}{\cellcolor{coolblue!40}Sim} \\ 
        \rowcolor{midgray}\cellcolor{darkgray}Backbone & \cellcolor{coolblue!30}SCT & \cellcolor{coolblue!30}SPL & \cellcolor{coolblue!30}SR \\ \myrule
        \cellcolor{lightgray}DinoV2   & \textbf{33.6} & \textbf{76.9} & \textbf{90.2} \\
        \cellcolor{lightgray}VC-1     & 31.8 & 75.1 & 89.0  \\
        \cellcolor{lightgray}Dune     & 32.1 & 73.7 & 87.6  \\
        \cellcolor{lightgray}AM-Radio & 31.1 & 73.9 & 88.1  \\
        \cellcolor{lightgray}DinoV3   & 0.00 & 0.00 & 0.00  \\ \myrule
    \end{tabularx}
    \caption{\label{tab:sap}\textbf{Self-Attention Pooling} -- as an ablation on \textit{DinoV3}. The failure of this projector indicates that complexity of the attention pooling is (in part) responsible for the failure of \textit{DinoV3}.}

\end{wrapfigure}

We conducted an additional experiment with a different projector preserving spatiality (similar to PAP and LPR), yet with a expressivity power comparable to CAP, ie. self-attention pooling (SAP) where the value projection matrix $\bW_v$ projects the path embedding to a smaller dimension (4 scalars per patch):
$$\bmf_t = \{\sigma\left(\bF_t \bW_{q}\cdot (\bF_t \bW_{k})^\top\right) \bW_{v}\bF_t$$
where $\bW_{v}\in \RR^{E\times 4}$. We report the metrics obtained with this variant in table \ref{tab:sap}, which failed again using \textit{DinoV3} as visual encoder. It indicates that compression and spatiality are not sufficient for enabling \textit{DinoV3}. The complexity (i.e. expressivity) of the projector is also at stake, since more expressive projectors seem to struggle with this vision encoder.

\end{document}